\documentclass[sigconf,10pt]{acmart}

\setcopyright{acmlicensed}
\copyrightyear{2026}
\acmYear{2026}
\acmDOI{XXXXXXX.XXXXXXX}
\acmISBN{978-1-4503-XXXX-X/2026/04}

\hypersetup{
    linkcolor=blue,   
    citecolor=red,    
    urlcolor=magenta  
}
\usepackage{algorithm}
\usepackage{appendix}
\usepackage{tikz}
\usepackage{amsthm}
\usepackage{listings}
\usepackage{bm}
\usepackage{xspace}
\usepackage{comment}
\usepackage{subcaption}
\usepackage{svg}
\usepackage[normalem]{ulem} 
\usepackage{sidecap}

\usepackage[most]{tcolorbox}

\newcounter{finding}
\newtcolorbox{findingbox}{
    colback=gray!5,
    colframe=black!55,
    boxrule=0.3pt,
    arc=1pt,
    left=3pt,
    right=3pt,
    top=1pt,
    bottom=1pt,
    before skip=0.3em,
    after skip=0.3em,
    before upper={\textbf{Finding~\refstepcounter{finding}\thefinding:}\ }
}

\usepackage[noend]{algpseudocode}
\usepackage{multirow}

\usepackage{enumitem}
\setlist{topsep=0pt, nosep, leftmargin=*}

\DeclareGraphicsExtensions{.png,.pdf}

\newcommand{\company}{CompanyX\xspace}

\DeclareSymbolFont{cmletters}{OML}{cmm}{m}{it}

  \usepackage{microtype}              
  \let\origunderscore\_
  \renewcommand{\_}{\discretionary{}{}{}\origunderscore\discretionary{}{}{}}

\begin{document}

\title{A Year in LLM Serving: Workload Evolution, Caching and Load-Balancing}

\author{William Nixon}
\affiliation{%
\institution{University of Chicago}
\country{}
}

\affiliation{%
\institution{Harvard University}
\country{}
}

\author{Jon Durbin}
\affiliation{%
\institution{Chutes}
\country{}
}

\author{Florian Standhartinger}
\affiliation{%
\institution{Chutes}
\country{}
}

\author{Haryadi S. Gunawi}
\affiliation{%
\institution{University of Chicago}
\country{}
}

\author{Juncheng Yang}
\affiliation{%
\institution{Harvard University}
\country{}
}

\renewcommand{\shortauthors}{Nixon et al.}

\begin{abstract}
Large Language Model (LLM) serving has become a critical
cloud workload, and realistic traces are essential for motivating
and benchmarking serving systems. However, existing LLM serving
workload studies remain limited in scale and scope. They often
observe short time periods and provide limited visibility into how
users interact with models in production. As a result, they do not
fully capture how LLM serving workloads evolve over time or how
user-model interactions shape production traffic.

In this work, we further the understanding of
real-world LLM serving workloads through both a global
characterization and a longitudinal study of a one-year production
trace from \company. Unlike prior studies, our trace captures
full production behavior across many models and users, including
both popular and long-tail models. We analyze the workload from
aggregate, temporal, model-level, and user-level perspectives,
revealing workload evolution and user-model structure that are
typically hidden behind aggregate views. 
To support future research, we publicly release the full one-year trace, enabling downstream studies of production behavior without relying on sampled or synthetically generated workloads. The trace is available at \url{https://github.com/HarvardMadSys/chutes_workload}.
\end{abstract}



\maketitle

\section{Introduction}
\label{sec:intro}

Large language models (LLMs) have become a core online service primitive, powering conversational assistants, code-generation tools, agentic workflows, retrieval-augmented generation pipelines, and programmatic API workloads. Serving these applications at low latency and high throughput remains challenging because autoregressive LLM inference is both computationally intensive and highly workload-dependent. Request arrival patterns determine batching efficiency and queueing delay; prompt and response lengths shape the balance between prefill and decode computation; model popularity influences placement, routing, and load balancing; and user behavior induces temporal locality, prefix reuse, and repeated-access patterns that affect KV-cache efficiency~\cite{yu2022orca,kwon2023vllm,zhong2024distserve,zheng2024sglang}. Consequently, realistic production traces are critical for understanding LLM serving behavior and for evaluating serving-system designs under representative operating conditions~\cite{wang2025burstgpt, wang2025kvcache}.

Recent studies have characterized LLM serving workloads and have shown that production traces differ substantially from synthetic arrival processes, static prompt datasets, and conventional cloud-service traces. These studies have provided important insights into bursty request arrivals, prompt and response length distributions, workload generation, and KV-cache reuse behavior~\cite{wang2025burstgpt, wang2025kvcache,xiang_servegen_2025,qiu_modserve_2025}. However, important gaps remain in our understanding of production LLM serving workloads. First, many existing studies analyze relatively short observation windows, limiting our ability to study how workloads evolve over longer time scales. Second, user--model interactions remain underexplored, even though modern LLM platforms serve many models with highly skewed popularity and support users with diverse access patterns. Third, existing public workloads are often sampled, anonymized, or summarized, which constrains direct replay and limits downstream analyses of production behavior. These limitations motivate a longitudinal study of LLM serving workloads that jointly examines workload evolution, caching opportunities, and load-balancing implications.

\begin{table}[t]
\scriptsize
\centering
\setlength{\tabcolsep}{4.5pt}
\renewcommand{\arraystretch}{1.08}
\caption{
Comparison with prior interaction datasets and production LLM workload studies.
\vspace{-0.6em}
}
\label{tab:workload_comparison}
\begin{tabular}{@{}lccccc@{}}
\toprule
\textbf{Work} &
\textbf{Duration} &
\textbf{\# Requests} &
\textbf{\# Models} &
\textbf{User} &
\textbf{Instance} \\
\midrule

ServeGen~\cite{xiang_servegen_2025} &
4 mo. &
3.54B &
12 &
\checkmark &
-- \\

BurstGPT~\cite{wang_burstgpt_2025} &
4 mo. &
5.29M &
2 &
-- &
-- \\

WildChat~\cite{zhao_wildchat_2024} &
13 mo. &
1.0M convs. &
2 &
\checkmark &
-- \\

LMSYS-Chat-1M~\cite{zheng_lmsys-chat-1m_2024} &
4 mo. &
1.0M convs. &
25 &
-- &
-- \\

ModServe~\cite{qiu_modserve_2025} &
1 wk &
-- &
-- &
-- &
-- \\

Qwen-Bailian~\cite{wang_kvcache_2025} &
2 h &
-- &
1 &
-- &
-- \\

Mooncake~\cite{qin_mooncake_2025} &
1 h &
35.6K &
-- &
-- &
-- \\

\textbf{Ours} &
\textbf{12 mo.} &
\textbf{6.12B} &
\textbf{9,174} &
\checkmark &
\checkmark \\

\bottomrule
\end{tabular}

\vspace{0.25em}
\raggedright
\footnotesize
\vspace{-1.5em}
\end{table}
\vspace{-0.25em}
 
\textit{To bridge this gap, we collect, characterize, and release a one-year production LLM serving trace} from \company. To the best of our knowledge, this is the first public LLM serving trace that provides a long observation window, unsampled request-level records, broad coverage across both popular and long-tail models, and anonymized user and serving-instance identifiers. This trace enables analyses that are difficult to perform from aggregate statistics alone. We study how LLM serving demand evolves over time, how production usage differs across models, and how users interact with those models. The request-level unsampled information allows for detailed cache analysis and simulation. We find that cache-eviction algorithms designed for web and storage workloads can underperform simple FIFO and LRU policies on LLM prefix-cache workloads. We also identify a fundamental tension in LLM serving: load balancing spreads requests across instances to improve utilization, but doing so can reduce prefix-cache locality, lowering cache hit ratio and increasing cache replication.

Our key contributions are as follows:
\begin{enumerate}
    \item \textbf{A year-long production LLM serving trace to support open research.}
    We collect, characterize, and release~\footnote{The trace is available at \url{https://github.com/HarvardMadSys/chutes_workload}.} a one-year, unsampled, request-level production trace from \company, covering billions of requests, thousands of models, anonymized users, serving instances, token counts, latency, and prefix-cache reuse.

    \item \textbf{Characterization of workload evolution.}
    We show that LLM serving workloads are non-stationary over yearly timescales. Request characteristics shift toward longer inputs and shorter outputs, changing the relative pressure on prefill and decode and making short measurement windows insufficient for capacity planning.

    \item \textbf{Long-tail model behavior and serving opportunities.}
    We find that many models receive sparse traffic with long inter-arrival times, while consecutive inter-arrival intervals often remain correlated. This structure creates opportunities to colocate or multiplex low-traffic models on shared GPUs without treating each model as an independent always-on service.

    \item \textbf{User--model interaction structure.}
    We show that users interact with models in heterogeneous ways: some repeatedly use a single model, while others switch across many models. These patterns are hidden in aggregate statistics and suggest opportunities for user-aware scheduling, placement, and caching.

    \item \textbf{Prefix-cache behavior in production.}
    We characterize prefix-cache reuse at request level and find that token hit ratios are often bimodal, with most requests seeing either near-zero or near-complete reuse. Reuse is also temporally concentrated, with much of it occurring within short inter-arrival windows. In simulation, state-of-the-art cache policies designed for web and storage workloads often underperform simple FIFO and LRU, motivating new cache-management policies for LLM prefix workloads.

    \item \textbf{Tension between caching and load balancing.}
    We identify a fundamental tradeoff in LLM serving: routing related requests to the same host improves prefix-cache locality and hit ratio, but also increases load imbalance and cache replication. This tension complicates scheduler design and motivates joint treatment of routing, placement, and caching.

\end{enumerate}

\section{Background}
\label{sec:bg}

\subsection{LLM Inference}

LLM inference consists of two main phases: \emph{prefill} and \emph{decode}. In the prefill phase, the serving system processes the input prompt and builds the request's key-value (KV) cache. In the decode phase, the model generates output tokens autoregressively, using the KV cache from previous tokens. Thus, the cost of a request depends on both input tokens and output tokens: input length mainly affects prefill cost, while output length affects how long the request occupies decode iterations.

This distinction matters for workload analysis. Request arrivals affect batching and queueing, input tokens affect prefill pressure, output tokens affect decode pressure, and cached tokens indicate how much prompt work can be avoided through prefix-cache reuse. Therefore, our trace focuses on request-level timing, model identifiers, user identifiers, token counts, latency, time-to-first-token, and cached-token information.

\subsection{LLM Workload}
\label{sec:background_workload}
Understanding LLM workloads is essential for designing practical serving systems. Properties such as request arrivals, token lengths, and user behavior directly shape decisions around batching, scheduling, caching, routing, and autoscaling. Motivated by this need, prior work has studied and released LLM workloads from several complementary angles. Public multi-turn datasets capture real conversational behavior~\cite{zhao_wildchat_2024, zheng_lmsys-chat-1m_2024}, while production studies examine specific serving phenomena such as KV-cache reuse and inference-system behavior~\cite{wang_kvcache_2025, qin_mooncake_2025, qiu_modserve_2025}. Other work focuses more directly on broader production workload characterization, including request dynamics and user-level structure~\cite{wang_burstgpt_2025, xiang_servegen_2025}. Table~\ref{tab:workload_comparison} summarizes prior workload-analysis studies and the traces or public workload artifacts they release.

However, existing public artifacts remain limited for downstream workload analysis. Some traces are sampled, some remove fields such as user identifiers even when those fields are used in the original study, and others release synthetic generators rather than full request-level production traces. These limitations make it difficult to independently study long-term workload evolution,  user--model interactions, or faithful inter-arrival behavior due to sampling.

To address this gap, we release and analyze a one-year production
trace containing 6.12 billion requests across 9,174 models. The trace
retains anonymized user identifiers, serving-instance identifiers,
and serving-side signals such as TTFT and cached-token counts.
This combination enables a longitudinal and fine-grained study of
production LLM serving: we examine how workloads evolve over
time, how users and models interact, how prefix reuse emerges,
and how these behaviors shape routing and load-balancing decisions.

\section{Production Trace Analysis}
\label{sec:traceoverview}

\begin{table}[t]
    \scriptsize
    \centering
    \setlength{\tabcolsep}{4pt}
    \renewcommand{\arraystretch}{1.06}
    \caption{Summary of the \company production trace.}
    \label{tab:chutes_trace_overview}

    \begin{tabular}{@{}lr@{}}
    \toprule
        \textbf{Metric} & \textbf{Value} \\
    \midrule
        Time span & 1 year (2025-04-11 to 2026-04-12) \\
        Requests & 6,122,413,756 \\
        Input tokens & 35,795,761 million \\
        Output tokens & 2,522,394 million \\
        Users & 314,970 \\
        Models & 9,174 (3,922 public, 5,252 private) \\
        Serving instances & 875,921 \\
    \bottomrule
    \end{tabular}
    \vspace{-2em}
\end{table}

We analyze a one-year production trace from \company, a serverless LLM inference platform that serves both platform-hosted and user-deployed models. Table~\ref{tab:chutes_trace_overview} summarizes the trace scale: 6.12 billion requests from 314,970 users across 9,174 models. Of these models, 3,922 are public and 5,252 are private. Public models can be invoked by any user and include both user-published models and platform-hosted models such as DeepSeek, MiniMax, and Kimi; Private models are custom models deployed by individual users for restricted access, including user-fine-tuned variants. 

Table ~\ref{tab:chutes_trace_fields} lists the fields available in the trace. For each request, we record its timestamp, endpoint, model, user, and input/output token counts. We also collect serving-side execution information, including the instance that handled the request, completion latency, time to first token (TTFT), and cached-token counts. These signals let us study production behavior including prefix-cache reuse and instance-level load balancing.

Unless otherwise noted, our analysis uses the full year of text-based API requests, including both streaming and non-streaming chat and completion endpoints. To provide representative examples, we highlight three high-traffic models with distinct workload profiles: DeepSeek-V3.2~\cite{deepseek-ai_deepseek-v3_2025}, which is primarily used for chat and role-playing; DeepSeek-R1~\cite{deepseek-ai_deepseek-r1_2025}, which generates long reasoning outputs; and MiniMax-M2.5~\cite{minimax_m25_2026}, which is heavily used for coding and agentic workloads. Together, these models highlight several prominent workload classes and illustrate how production behavior varies across model types.

The remainder of the paper builds on this trace in stages. We first characterize aggregate workload structure in this section, then examine user and model level behavior (\autoref{sec:users_models}), study how the workload evolves over the year (\autoref{sec:evolution}), and analyze prefix caching in production (\autoref{sec:caching}). Finally, we use the trace to study the tension between load balancing and prefix-cache locality: spreading requests improves balance, while routing related requests to the same instance improves cache reuse.

\begin{table}[t]
    \scriptsize
    \centering
    \setlength{\tabcolsep}{4pt}
    \renewcommand{\arraystretch}{1.06}
    \caption{Key fields available in the \company production trace.}
    \label{tab:chutes_trace_fields}

    \begin{tabular}{@{}llll@{}}
    \toprule
        \textbf{Field} & \textbf{Description} &
        \textbf{Field} & \textbf{Description} \\
    \midrule
        \texttt{request\_id}    & Request ID &
        \texttt{endpoint}       & API endpoint \\

        \texttt{model}          & Model identifier &
        \texttt{user\_id}       & Anonymized user ID \\

        \texttt{instance\_id}   & Serving instance &
        \texttt{started\_at}    & Start timestamp \\

        \texttt{completed\_at}  & Completion timestamp &
        \texttt{input\_token}   & Input token count \\

        \texttt{output\_token}  & Output token count &
        \texttt{ttft}           & Time to first token \\

        \texttt{cached\_tokens} & Prefix-cached tokens &
        & \\
    \bottomrule
    \end{tabular}
\end{table}



\begin{figure*}[t]
    \centering
    \begin{subfigure}[t]{0.19\textwidth}
        \centering
        \includegraphics[width=\linewidth]{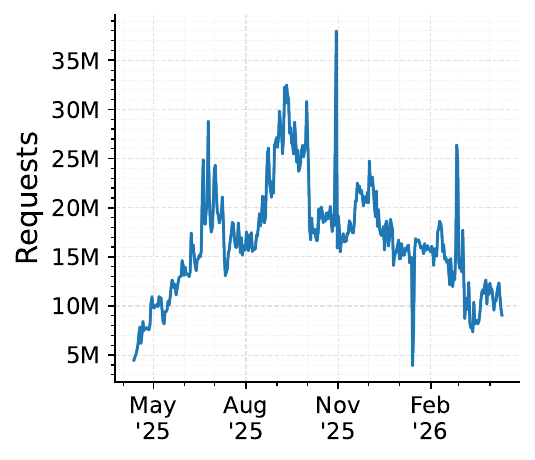}
        \caption{Daily request rate}
        \label{fig:request_rate_over_time}
    \end{subfigure}
    \hfill
    \begin{subfigure}[t]{0.19\textwidth}
        \centering
        \includegraphics[width=\linewidth]{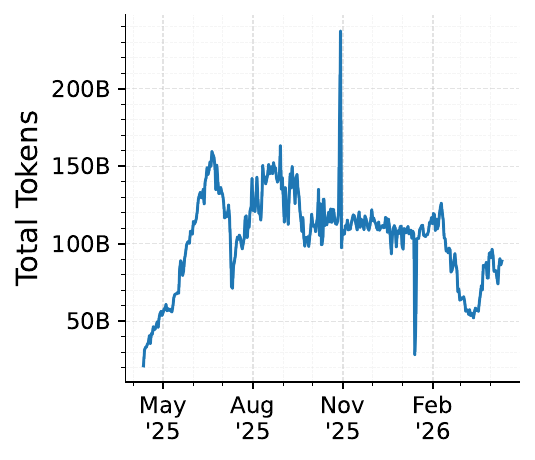}
        \caption{Daily token rate}
        \label{fig:todo_active_users_over_time}
    \end{subfigure}
    \hfill
    \begin{subfigure}[t]{0.19\textwidth}
        \centering
        \includegraphics[width=\linewidth]{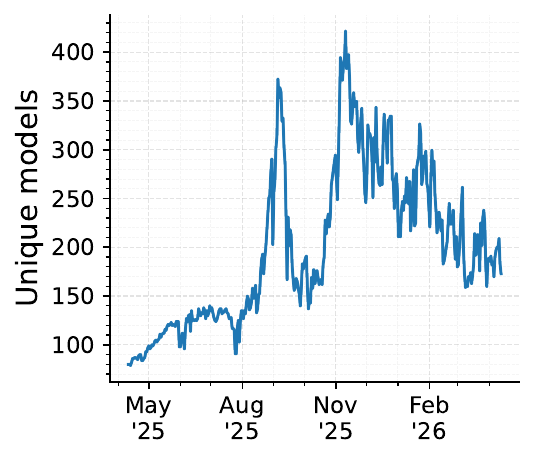}
        \caption{Daily active models}
        \label{fig:active_models_over_time}
    \end{subfigure}
    \hfill
    \begin{subfigure}[t]{0.19\textwidth}
        \centering
        \includegraphics[width=\linewidth]{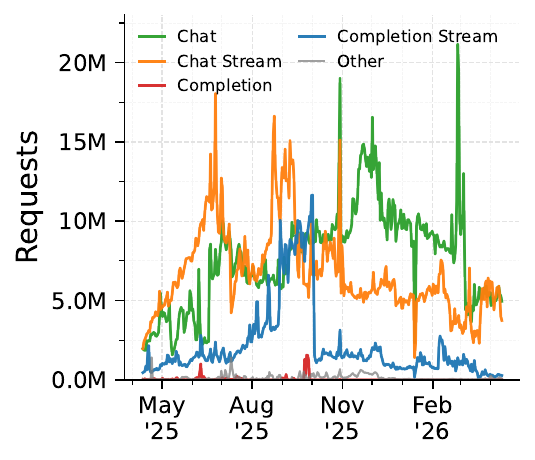}
        \caption{Daily API calls}
        \label{fig:api_call_over_time}
    \end{subfigure}
    \hfill
    \begin{subfigure}[t]{0.19\textwidth}
        \centering
        \includegraphics[width=\linewidth]{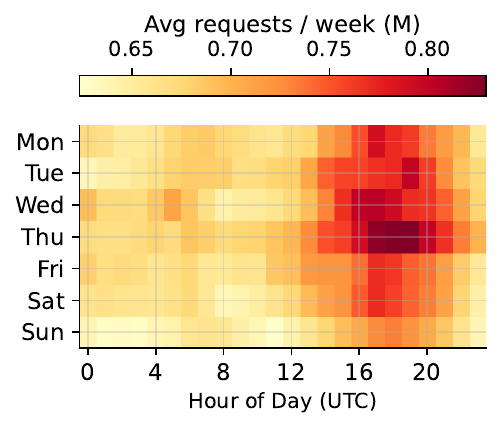}
        \caption{Request heatmap}
        \label{fig:hour_of_week_heatmap}
    \end{subfigure}
    \vspace{-2mm}
    \caption{Daily request volume, token volume, active models, and API mix evolve differently over the year, showing that production load is non-stationary and cannot be summarized by request count alone.}
    \label{fig:workload_temporal_structure}
\end{figure*}

\subsection{Workload Overview}
\label{sec:workload_overview}

\paragraph{Aggregate demand is non-stationary.} Production demand changes over both long and short timescales. Figure~\ref{fig:request_rate_over_time} shows that daily request volume rises through mid-2025, peaks around late summer and fall, and then declines into early 2026. Figure~\ref{fig:todo_active_users_over_time} shows that daily token volume follows a similar but not identical trend: it grows quickly early in the trace, remains elevated for much of the year, and drops more sharply near the end. Similar request volumes can therefore represent different amounts of serving work.

\paragraph{Active models and API mix shift over the year.} The workload also changes in breadth and usage mix. Figure~\ref{fig:active_models_over_time} shows that the number of active models per day grows from fewer than 100 early in the trace to more than 400 at its peak, before settling at a lower but still elevated level. Much of this fluctuation comes from user-deployed private models, whose availability and usage can change quickly. Over the same period, Figure~\ref{fig:api_call_over_time} shows that the API mix shifts: \textit{chat}-based traffic becomes more prominent, while \textit{completion}-based traffic declines, and streaming calls also become less dominant, with non-streaming functions accounting for a larger share later in the trace. Thus, production demand evolves not only in volume, but also in which models are active and how the platform is used.

\paragraph{Weekly and diurnal patterns persist throughout.} These long-term shifts coexist with a recurring weekly and diurnal pattern. Figure~\ref{fig:hour_of_week_heatmap} shows that traffic is highest during UTC afternoon and evening hours, especially on weekdays, and lower during overnight hours and weekends. Taken together, Figure~\ref{fig:workload_temporal_structure} shows that LLM serving demand cannot be summarized by a fixed request rate or a single growth trend. Realistic workload analysis must account for changing token load, model availability, API usage, and recurring weekly seasonality.

\subsection{Token Shape}
\label{sec:token_shape}

\begin{figure}[t]
    \centering
    \begin{subfigure}[t]{0.32\linewidth}
        \centering
        \includegraphics[width=\linewidth]{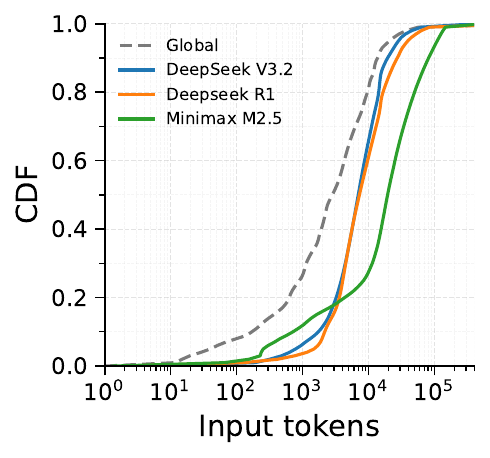}
        \caption{Input token}
        \label{fig:highlight_input_cdf}
    \end{subfigure}
    \begin{subfigure}[t]{0.32\linewidth}
        \centering
        \includegraphics[width=\linewidth]{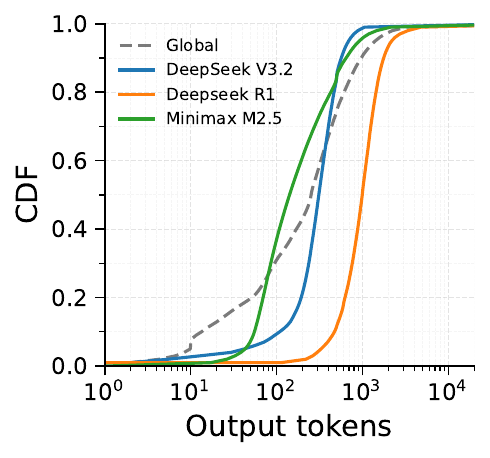}
        \caption{Output token}
        \label{fig:highlight_output_cdf}
    \end{subfigure}
    \begin{subfigure}[t]{0.32\linewidth}
        \centering
        \includegraphics[width=\linewidth]{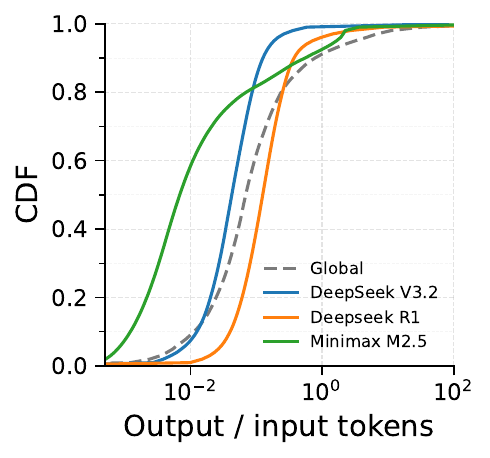}
        \caption{O/I ratio}
        \label{fig:highlight_ratio_cdf}
    \end{subfigure}
    \vspace{-0.5em}
    \caption{Request-level token-length distributions for three representative models against the full population. The models have distinctive characteristics.}
    \vspace{-1em}
    \label{fig:highlight_token_shape}
\end{figure}

\paragraph{Requests are often prompt-heavy.} Requests span a wide range of token lengths. Figure~\ref{fig:highlight_input_cdf} shows that input sizes vary from roughly \(10^2\) to \(10^5\) tokens, with a heavy tail toward very long prompts (not shown). Figure~\ref{fig:highlight_output_cdf} shows that output sizes are more concentrated, mostly between \(10^1\) and \(10^3\) tokens. Figure~\ref{fig:highlight_ratio_cdf} confirms that most requests generate fewer output tokens than input tokens. Together, these distributions show that the workload is often prompt-heavy, so prefill remains an important part of serving cost even when decode dominates end-to-end runtime.

\paragraph{Highlighted models show distinct usage profiles.} The model-level curves reveal that this global picture combines distinct usage profiles. MiniMax-M2.5 has longer inputs, shorter outputs, and the lowest output-to-input ratio, consistent with coding / agentic workloads that process large contexts but return relatively short responses. DeepSeek~R1 shows the opposite tendency as it has heavier output behavior and the highest output-to-input ratio, consistent with reasoning workloads that generate longer completions. DeepSeek~V3.2 lies closer to the global distribution, matching its more typical chat and role-play usage. Thus, models with similar request counts can impose very different serving pressure depending on whether their traffic is dominated by long prompts, long generations, or a more balanced mix of both.

\subsection{Latency}
\label{sec:latency_overview}

\begin{figure}[t]
    \centering
    \begin{subfigure}[t]{0.49\linewidth}
        \centering
        \includegraphics[width=\linewidth]{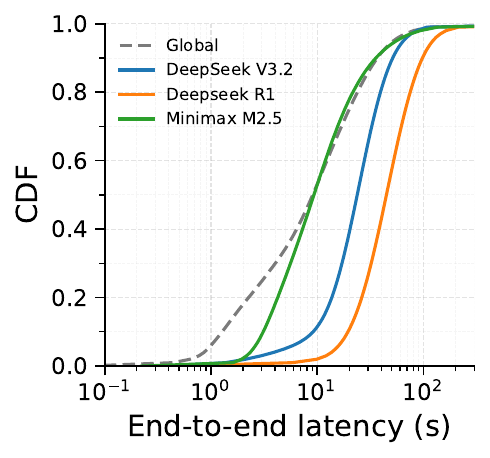}
        \caption{Latency CDFs}
        \label{fig:latency_cdf_global}
    \end{subfigure}
    \hfill
    \begin{subfigure}[t]{0.49\linewidth}
        \centering
        \includegraphics[width=\linewidth]{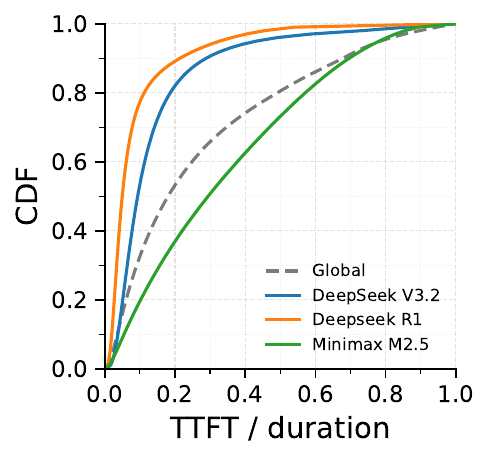}
        \caption{Fraction of TTFT}
        \label{fig:todo_ttft_cache_bucket}
    \end{subfigure}
    \vspace{-2mm}
    \caption{Latency behavior of the aggregate workload and three representative models. Models spend most proportion of their time doing decode.}
    \vspace{-3mm}
    \label{fig:cache_latency_pair}
\end{figure}
\paragraph{End-to-end latency is dominated by decode.} We decompose request latency into time to first token (TTFT), which captures the prefill-side delay before generation, and the remaining completion time. Figure~\ref{fig:latency_cdf_global} shows that end-to-end latency spans several orders of magnitude, from sub-second requests to requests lasting tens of seconds. Figure~\ref{fig:todo_ttft_cache_bucket} shows that TTFT is usually only a fraction of total request duration, indicating that end-to-end latency is dominated by decode.

\paragraph{Models can be latency-bound for different reasons.} The three highlighted models reveal distinct latency profiles. DeepSeek~R1 has the longest end-to-end latency, consistent with its tendency to generate long outputs. MiniMax-M2.5 completes faster overall, but a larger share of its latency comes from TTFT, reflecting its input-heavy workload. DeepSeek~V3.2 lies between these two cases. Thus, models may be decode-heavy due to long outputs or prefill-heavy due to long prompts, motivating phase-specific rather than one-size-fits-all optimization.

\begin{figure}[t]
    \centering
    \begin{subfigure}[t]{0.32\linewidth}
        \centering
        \includegraphics[width=\linewidth]{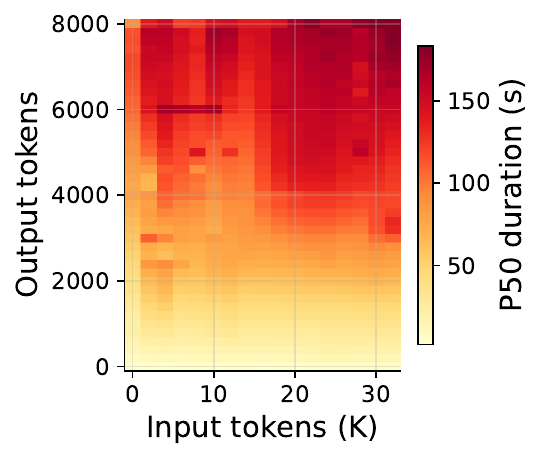}
        \caption{Output vs. Duration}
        \label{fig:todo_ttft_sequence_length}
    \end{subfigure}
    \hfill
    \begin{subfigure}[t]{0.32\linewidth}
        \centering
        \includegraphics[width=\linewidth]{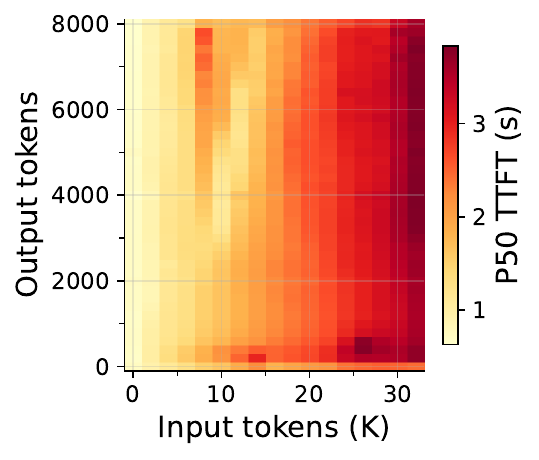}
        \caption{Input vs. TTFT}
        \label{fig:todo_ttft_request_rate}
    \end{subfigure}
    \vspace{-2mm}
    \begin{subfigure}[t]{0.32\linewidth}
        \centering
        \includegraphics[width=\linewidth]{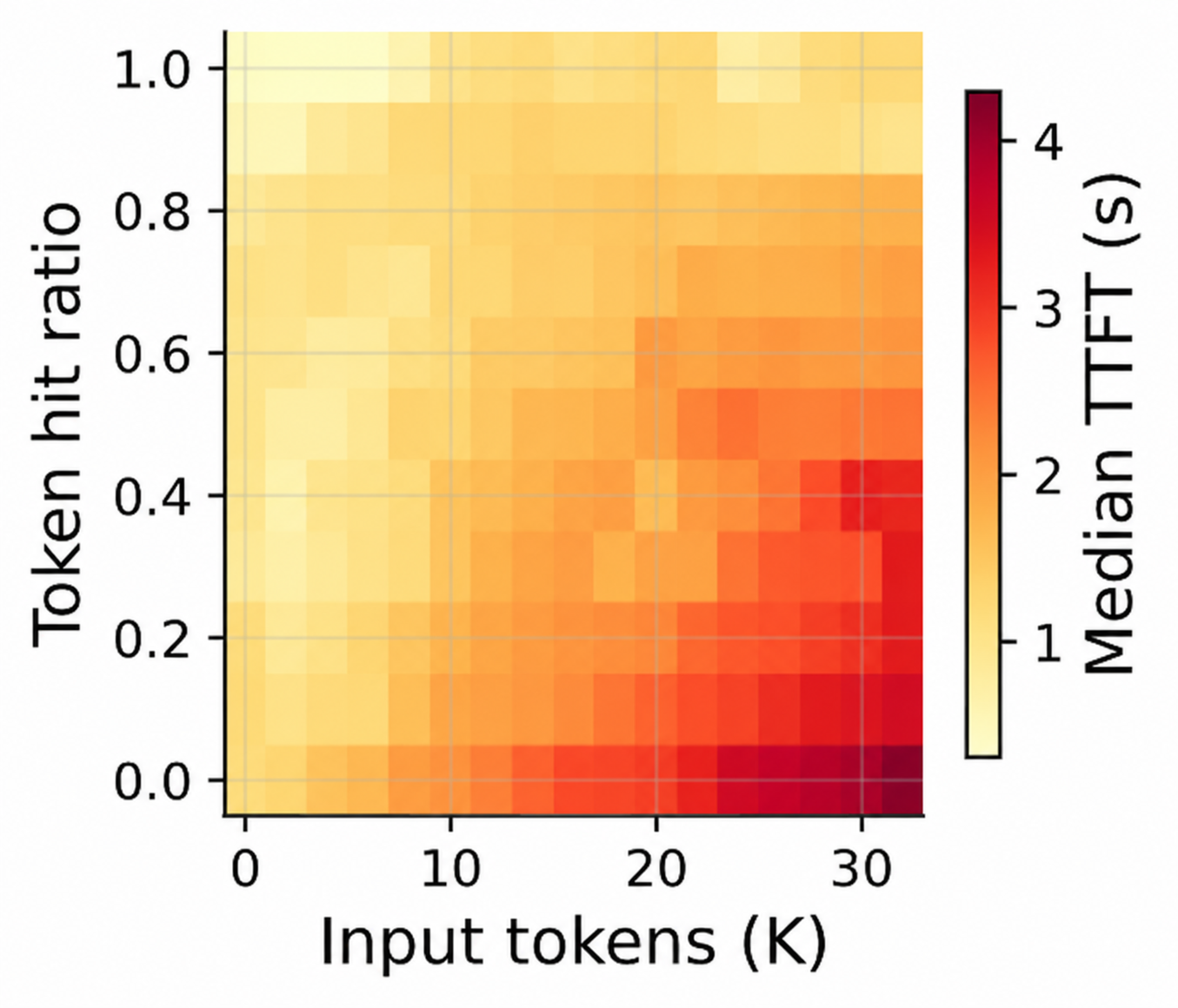}
        \caption{TTFT vs. hit ratio}
        \label{fig:todo_ttft_cache_effect}
    \end{subfigure}
    \caption{Median latency binned by request shape. Duration is governed by output length, TTFT by input length, and caching cuts TTFT most aggressively on long-input requests.}
    \vspace{-4mm}
    \label{fig:todo_ttft_diagnostics_pair}
\end{figure}

\paragraph{Output length drives duration; input length drives TTFT.} Binning requests by token shape makes these latency drivers explicit. Figure~\ref{fig:todo_ttft_sequence_length} shows that median end-to-end duration increases proportionally to output length, getting barely affected by input token length since completion time is decode dominated, and figure~\ref{fig:todo_ttft_request_rate} shows TTFT grows primarily with input length. Figure~\ref{fig:todo_ttft_cache_effect} connects this prefill cost to prefix reuse. For requests with similar input length, higher token hit ratios correspond to lower TTFT, and the reduction is most visible for long-input requests, where recomputing prefill would have been expensive. For short-input requests, even near-full cache hits remove little prefill work, so the TTFT benefit is small and contributes little to end-to-end latency. While caching may still save computation in these cases, but its latency impact is limited. 
\section{Users and Models}
\label{sec:users_models}

Aggregate workload statistics do not fully describe how individual users and models contribute to production traffic. We therefore examine the workload at finer granularity along two dimensions. First, we characterize the differences of users and models that may have similar request volumes (\autoref{sec:heterogeneity}). Second, we study their access behavior, including recurring usage patterns and arrival burstiness (\autoref{sec:access_patterns} and~\autoref{sec:burstiness}).

\subsection{Heterogeneity}
\label{sec:heterogeneity}
\begin{figure}[t]
    \centering
    \begin{subfigure}[t]{0.49\linewidth}
        \centering
        \includegraphics[width=\linewidth]{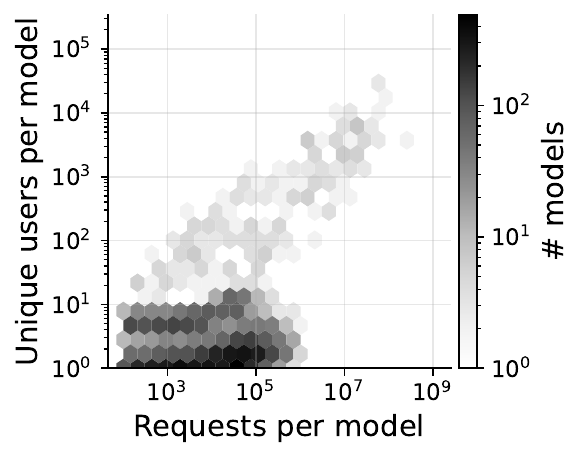}
        \caption{Models: requests vs.\ users}
        \label{fig:volume_breadth_models}
    \end{subfigure}
    \hfill
    \begin{subfigure}[t]{0.49\linewidth}
        \centering
        \includegraphics[width=\linewidth]{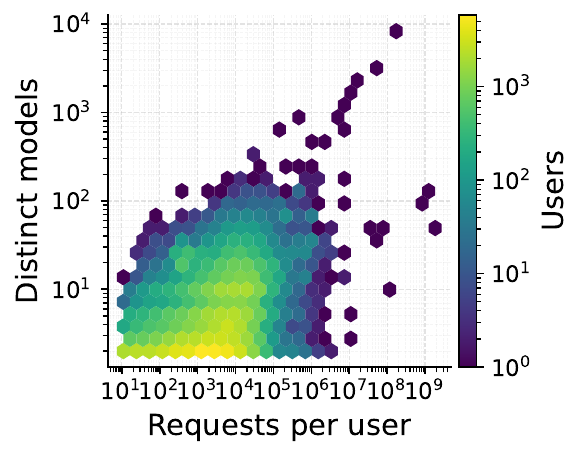}
        \caption{Users: requests vs.\ models}
        \label{fig:volume_breadth_users}
    \end{subfigure}
    \vspace{-2mm}
    \caption{Requests and model usage at the model and user level. Models and users with similar request volumes can differ substantially in access breadth.}
    \vspace{-4mm}
    \label{fig:volume_breadth}
\end{figure}

\paragraph{Models and users span wide ranges of request volume.} Request volume varies substantially across both models and users. Figure~\ref{fig:volume_breadth} relates request count to user--model interaction from both directions: for each model, we compare its request count with the number of distinct users it serves; for each user, we compare their request count with the number of distinct models they access.

On the model side, Figure~\ref{fig:volume_breadth_models} shows a weak positive trend with substantial spread. Models with more requests often serve more users, yet models with similar request counts can differ by orders of magnitude in user count. Most models, including many heavily used ones, receive the majority of their traffic from relatively few users. Some high-traffic models are used broadly, while others receive comparable traffic from a much smaller user base. This distinction matters for locality: repeated access from a concentrated set of users can create different cache-reuse opportunities than traffic spread across many independent users. The platform’s support for user-deployed models likely contributes to this spread, since private models may receive steady traffic from only a small, fixed group of users. 

Figure~\ref{fig:volume_breadth_users} shows that most users issue relatively few requests and access only a small number of models. However, a sparse tail of users sends many requests across a much broader model set. Most users therefore concentrate their activity on a few models, while these high-volume, broad-access power users may correspond to automated applications, proxy services, model routers, or agentic workflows rather than simple individual use.

\begin{figure}[t]
    \centering
    \begin{subfigure}[t]{0.49\linewidth}
        \centering
        \includegraphics[width=\linewidth]{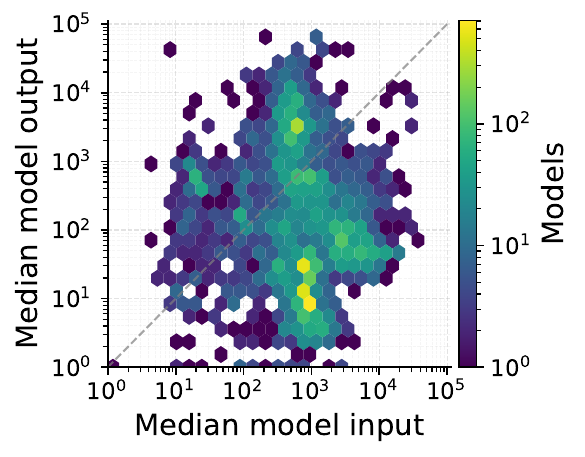}
        \caption{Models: input vs. output}
        \label{fig:token_shape_models}
    \end{subfigure}
    \hfill
    \begin{subfigure}[t]{0.49\linewidth}
        \centering
        \includegraphics[width=\linewidth]{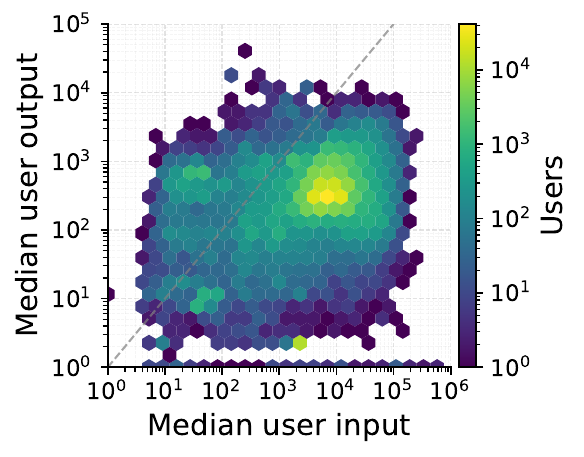}
        \caption{Users: input vs. output}
        \label{fig:token_shape_users}
    \end{subfigure}
    \vspace{-2mm}
    \caption{Most models and users have longer median inputs than outputs, but vary substantially across groups.}
    \label{fig:entity_token_shape}
\end{figure}

\paragraph{Token shape varies substantially across models and users.} Figure~\ref{fig:entity_token_shape} places each model or user by its median input and output length, with the dashed line marking equal size. Most points fall below the line, which means that most users and models tend to be more input heavy than output at the median. However, the spread is wide, showing that this average hides very different request shapes. The model view in Figure~\ref{fig:token_shape_models} shows two distinct regions. Some models are strongly input-heavy, receiving long prompts but producing shorter outputs, while others are output-heavy, generating responses that are large relative to their inputs. The user view in Figure~\ref{fig:token_shape_users} is more concentrated, but still spans a broad range. Some users consistently send short prompts, while others work with much larger contexts; output lengths also vary substantially. Taken together, request count and token shape describe different aspects of workload behavior. A model or user with a similar number of requests may still impose very different serving costs depending on whether their traffic is input-heavy, output-heavy, or more balanced. Together, these results show that models and users with similar traffic can still place very different demands on the system.

\begin{figure}[t]
    \centering
    \begin{subfigure}[t]{0.49\linewidth}
        \centering
        \includegraphics[
            width=\linewidth,
            trim=0 0 0 22,
            clip
        ]{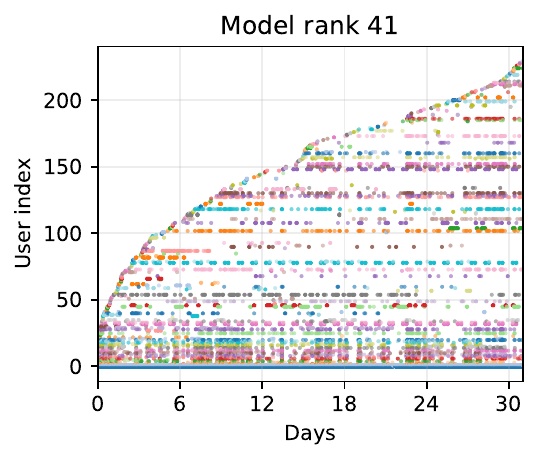}
        \caption{Periodic user access}
        \label{fig:model_access_power_user}
    \end{subfigure}
    \hfill
    \begin{subfigure}[t]{0.49\linewidth}
        \centering
        \includegraphics[
            width=\linewidth,
            trim=0 0 0 22,
            clip
        ]{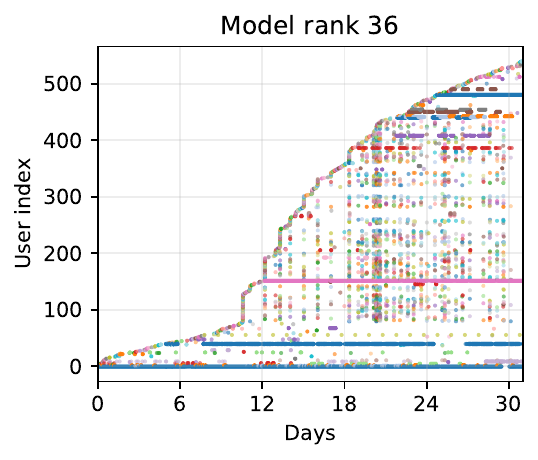}
        \caption{Correlated access}
        \label{fig:model_access_correlated}
    \end{subfigure}
    \vspace{-2mm}
    \caption{Thirty-day user-access rasters for two popular models. Models with similar popularity can exhibit different usage patterns from individual users.}
    \label{fig:model_access_patterns}
    \vspace{-1em}
\end{figure}

\subsection{Access Patterns}
\label{sec:access_patterns}

Aggregate request volume does not show how user--model interactions unfold over time. Users with similar request counts may access models in different ways, and popular models may receive traffic from different user patterns. We illustrate this with 30-day user-centric rasters show which models a user accesses over time and model-centric rasters show which users access a model over time.
\begin{figure}[t]
    \centering
    \begin{subfigure}[t]{0.49\linewidth}
        \centering
        \includegraphics[width=\linewidth, trim= 0 0 0 32, clip]{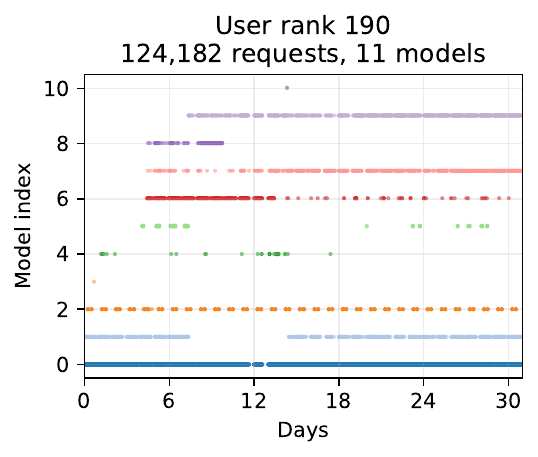}
        \caption{Periodic user usage}
        \label{fig:user_access_persistent}
    \end{subfigure}
    \hfill
    \begin{subfigure}[t]{0.49\linewidth}
        \centering
        \includegraphics[width=\linewidth, trim=0 0 0 32, clip]{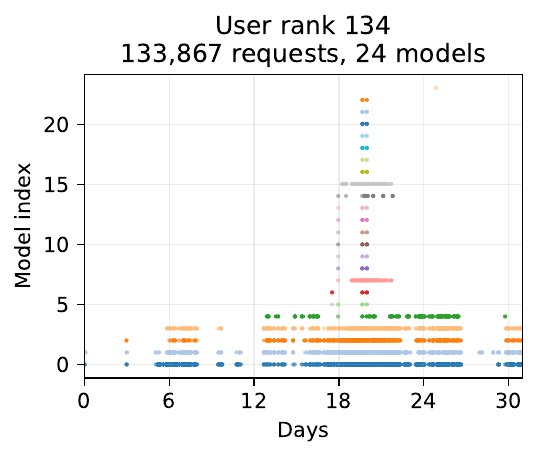}
        \caption{Model exploration}
        \label{fig:user_access_exploration}
    \end{subfigure}
    \vspace{-2mm}
    \caption{Thirty-day model-access rasters for two high-volume users. Users with similar request counts can show different access patterns.}
    \vspace{-5mm}
    \label{fig:user_access_patterns}
\end{figure}

\paragraph{Heavy users differ in how broadly they access models.} Figure~\ref{fig:user_access_patterns} shows two high-volume users with different access behavior. In Figure~\ref{fig:user_access_persistent}, the user repeatedly accesses a small and stable set of models across the 30-day window. Some models are used throughout most of the month, while others are accessed more periodically. Figure~\ref{fig:user_access_exploration} shows a different pattern: the user spends most of the month on a few models, but briefly accesses many additional models during a short period. These examples show that a single user may use multiple models concurrently, either steadily or in short bursts. Such behavior is plausible on this platform, where users can build model routers that send requests to different models based on their own criteria. The more periodic patterns reflect recurring daily activity or scheduled jobs.

\paragraph{Popular models can exhibit very different user-level access patterns.}
Figure~\ref{fig:model_access_patterns} compares thirty-day user-access rasters for two popular models. Figure~\ref{fig:model_access_power_user} also shows a largely \emph{periodic} pattern. Many users remain active across long portions of the window, forming horizontal bands that persist over days or weeks suggesting activity of power users. Several users issue requests repeatedly at regular intervals, while new users appear gradually throughout the month. Figure~\ref{fig:model_access_correlated} shows a more \emph{correlated} pattern. Activity is sparse early in the window, then expands sharply around the middle of the month as many users become active within the same period. The dense vertical clusters indicate that requests from different users arrive at similar times, rather than being spread independently across the window. One possible explanation is that related accounts or coordinated applications invoke the model in similar ways; for example, multiple accounts may be used to distribute traffic under platform limits.

\paragraph{Opportunity for pattern-aware serving.} User- and model-level behavior provides actionable signals for serving decisions. Mining block, job, and application level patterns has improved caching and scheduling policies, despite the difficulty of learning from such fine-grained signals~\cite{yang_mithril_2017, chung2020unearthing, ferguson2012jockey}. Block accesses span a vast space, while individual jobs reveal little about recurring behavior. User- and model-level histories provide a coarser and more stable view that may be easier to learn and exploit for caching, prefetching, routing, model placement, and capacity provisioning.

\begin{findingbox}
Users and models exhibit distinct temporal access patterns which can be used to inform serving decisions, including caching, routing, model placement, and capacity provisioning.
\end{findingbox}

\subsection{Burstiness}
\label{sec:burstiness}

Request arrivals may be steady or clustered into short bursts, which affects provisioning, scheduling, and placement. We characterize burstiness using the coefficient of variation (CV) of inter-arrival time (IAT): a CV greater than 1 indicates that request traffic is uneven.

\begin{figure}[t]
    \centering
    \begin{subfigure}[t]{0.49\linewidth}
        \centering
        \includegraphics[width=\linewidth]{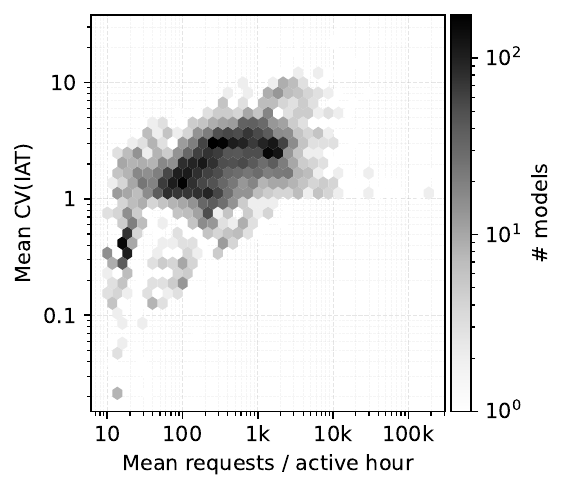}
        \caption{Popularity vs.\ burstiness}
        \label{fig:model_burstiness_density}
    \end{subfigure}
    \hfill
    \begin{subfigure}[t]{0.49\linewidth}
        \centering
        \includegraphics[
            width=\linewidth,
            trim=0 0 270pt 0,
            clip
        ]{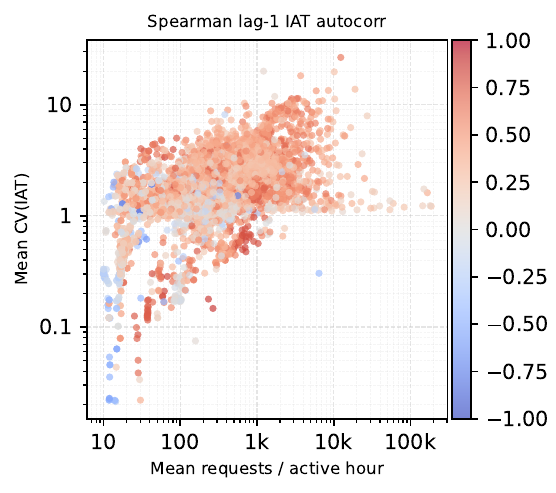}
        \caption{Lag-1 IAT autocorrelation}
        \label{fig:model_burstiness_autocorr}
    \end{subfigure}
    \vspace{-2mm}
    \caption{Model request load and burstiness. Most models are bursty and show positive correlation, indicating persisting high-activity and low-activity periods.}
    \label{fig:model_burstiness}
    \vspace{-1.5em}
\end{figure}

\paragraph{Most models are bursty and persistently so.} Figure~\ref{fig:model_burstiness_density} plots each model by its mean request rate during active hours and the minutely CV of its inter-arrival times. Most models lie above CV~$=1$, showing that their arrivals are bursty rather than evenly spaced. This pattern appears across both low- and high-rate models, so models with more traffic are not necessarily steadier. Figure~\ref{fig:model_burstiness_autocorr} shows that among these bursty models, many do show positive autocorrelation. This indicates that high-load periods tend to follow high-load periods, and low-load periods tend to persist. Model traffic often stays in an active or quiet state for some time instead of fluctuating randomly from one request to the next. 

\paragraph{Persistent bursts create colocation opportunities.} These patterns may matter for placement and autoscaling. Many lower-traffic models have busy and quiet periods that often persist for some time. This creates opportunities to colocate or multiplex models whose active windows do not overlap, rather than provisioning each sparse model as if it required dedicated capacity at all times. 

\begin{findingbox}
Model burstiness is often persistent and correlated. Busy periods and quiet periods tend to persist. This context may help autoscaling and colocation policies anticipate load.
\end{findingbox}

\section{One-Year Workload Evolution}
\label{sec:evolution}

In this section, we discuss how the workload evolved over the year, focusing on model usage (\autoref{sec:model_mix}) and token shape (\autoref{sec:token_shape_evolution}).

\subsection{Model Usage}
\label{sec:model_mix}

\begin{figure}[t]
    \centering
    \includegraphics[width=0.75\linewidth]{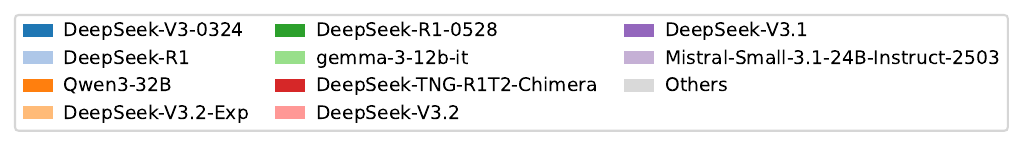}
    \vspace{-1mm}

    \begin{subfigure}[t]{0.49\linewidth}
        \centering
        \includegraphics[width=\linewidth]{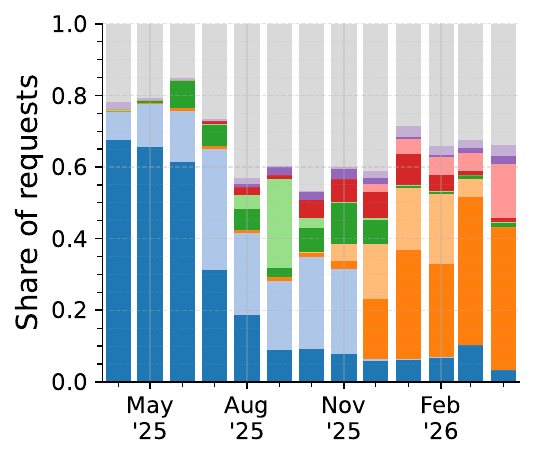}
        \caption{Request share}
        \label{fig:top10_model_monthly_request_share}
    \end{subfigure}
    \hfill
    \begin{subfigure}[t]{0.49\linewidth}
        \centering
        \includegraphics[width=\linewidth]{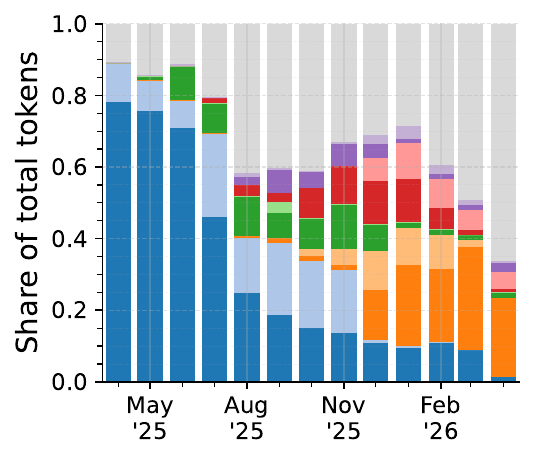}
        \caption{Token share}
        \label{fig:top10_model_monthly_token_share}
    \end{subfigure}

    \vspace{-2mm}
    \caption{Monthly model mix among the top models by request and token share. Dominant models and their token shares are becoming more diverse over time.}
    \label{fig:model_mix_over_time}
    \vspace{-4mm}
\end{figure}

\paragraph{Dominant models turn over.} The set of dominant models changes substantially over the year. Figure~\ref{fig:model_mix_over_time} shows the monthly share of traffic contributed by the top models, with the ``Others'' bucket capturing the remaining share. In the early months, traffic is dominated by DeepSeek models, such as DeepSeek-V3-0324 and DeepSeek-R1. Their shares decline over time, while newer models such as Qwen3-32B and DeepSeek-V3.2 become more prominent later on. The token-weighted view also shows the same trend. 

\paragraph{Traffic spreads across more models over time.} Model usage also becomes more diverse over time. The ``Others'' band grows in both panels, showing that a larger fraction of requests and tokens comes from the non-dominant models. As more models are released and adopted, traffic spreads across a wider range of models rather than remaining concentrated in a small group of dominant ones. This matters for system design as optimizations tuned to the most-used models in one period may become less suitable later since both the leading models and the broader model mix change.

\subsection{Token Shape}
\label{sec:token_shape_evolution}

\begin{figure}[t]
    \centering
    \begin{subfigure}[t]{0.49\linewidth}
        \centering
        \includegraphics[width=\linewidth]{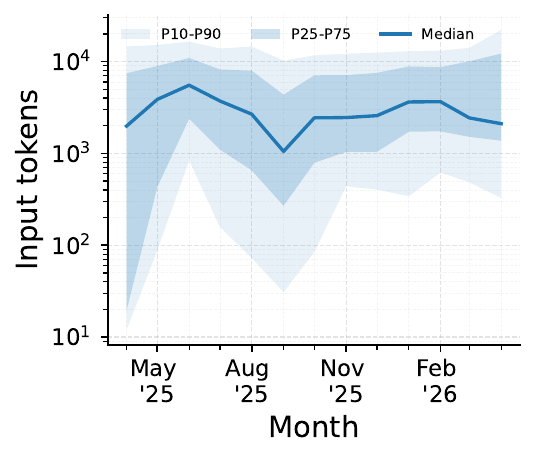}
        \caption{Monthly input intervals}
        \label{fig:monthly_input_token_intervals}
    \end{subfigure}
    \hfill
    \begin{subfigure}[t]{0.49\linewidth}
        \centering
        \includegraphics[width=\linewidth]{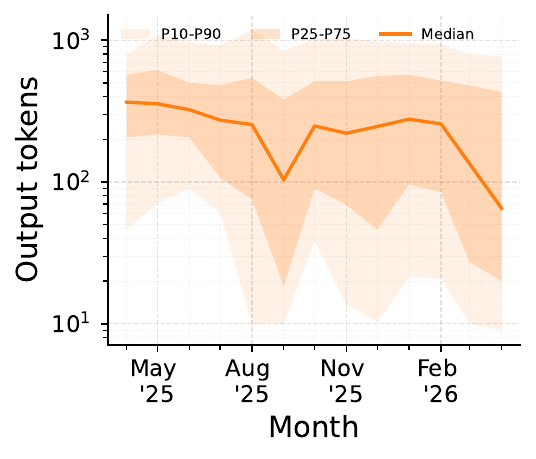}
        \caption{Monthly output intervals}
        \label{fig:monthly_output_token_intervals}
    \end{subfigure}
    \vspace{-2mm}
    \caption{Monthly input (a) and output (b) token-length bands (P10--P90, P25--P75, median). Global input tend to stay stable while output tends to drop over time.}
    \vspace{-4mm}
    \label{fig:todo_token_percentiles_over_time}
\end{figure}

Token lengths also change over the year, and input and output tokens evolve differently. We examine the aggregate monthly trend, then group users by their join time to understand whether these changes are associated with newer user cohorts.

\paragraph{Inputs stay stable while outputs shrink.} Figure~\ref{fig:todo_token_percentiles_over_time} shows the monthly distribution of request token lengths. Input lengths remain broadly stable over the year, with the monthly median staying in the thousands of tokens. Output lengths show a clearer downward trend, with the median falling from a few hundred tokens early in the trace to below one hundred tokens near the end. The wider bands toward the end of the trace also indicate greater variation in output lengths across requests. These aggregate bands are request-weighted, so users who contribute more traffic have a larger influence on the overall trend. To complement this view, we next group users by join time and examine the distribution of each user’s median input and output length within each cohort.

\begin{figure}[t]
    \centering
    \begin{subfigure}[t]{0.49\linewidth}
        \centering
        \includegraphics[
            width=\linewidth,
            trim=0 22 0 0,
            clip
        ]{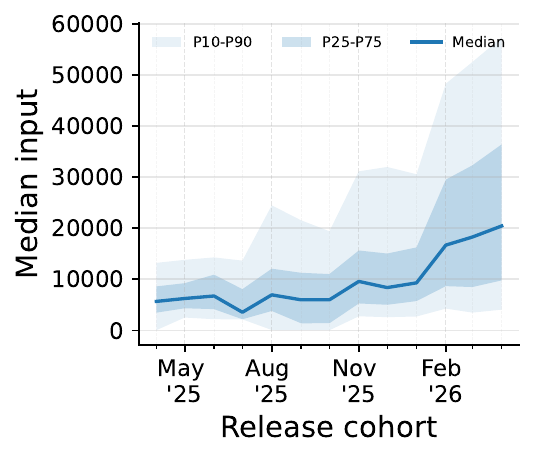}
        \caption{Median input by cohort}
        \label{fig:app_user_release_cohort_input}
    \end{subfigure}\hfill%
    \begin{subfigure}[t]{0.49\linewidth}
        \centering
        \includegraphics[
            width=\linewidth,
            trim=0 22 0 0,
            clip
        ]{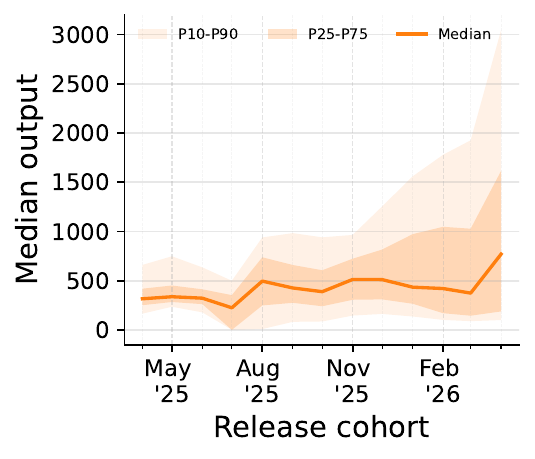}
        \caption{Median output by cohort}
        \label{fig:app_user_release_cohort_output}
    \end{subfigure}
    \vspace{-2mm}
    \caption{Per-user median input (a) and output (b) by release cohort, with P10--P90 and P25--P75 bands. Later user cohorts shift toward larger inputs and outputs}
    \label{fig:app_user_release_cohort_shape}
    \vspace{-6mm}
\end{figure}

\paragraph{Newer user cohorts are more input-heavy.} Cohort decomposition shows that newer user cohorts tend to be more input-heavy than earlier ones. Figure~\ref{fig:app_user_release_cohort_shape} shows that later cohorts have larger median input lengths, indicating that newer users typically submit longer prompts. Change in the median output side is less pronounced. However, the percentile bands also widen in both panels for later cohorts, showing greater variation across users in their typical input and output lengths. Overall, newer user cohorts tend to use longer inputs and exhibit more diverse token profiles than earlier cohorts.

\begin{findingbox}
Both model usage and token profiles change over time. Dominant models turn over, global outputs become shorter, and newer user show larger and more variable token lengths. 
\end{findingbox}

\begin{figure*}[t]
    \centering
    \begin{subfigure}[t]{0.23\textwidth}
        \centering
        \includegraphics[width=\linewidth]{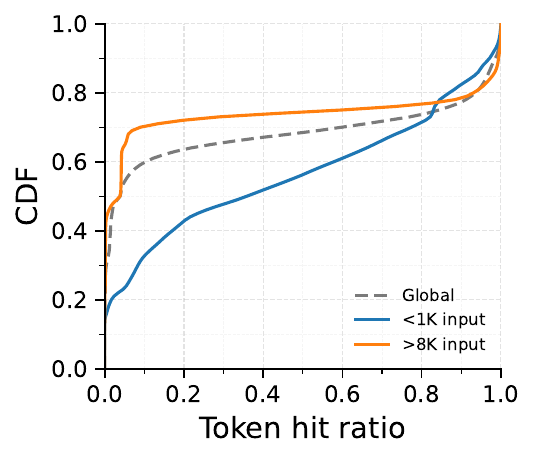}
        \caption{Prompt length and hit ratio}
        \label{fig:request_cache_fraction_cdf}
    \end{subfigure}\hfill
    \begin{subfigure}[t]{0.23\textwidth}
        \centering
        \includegraphics[width=\linewidth]{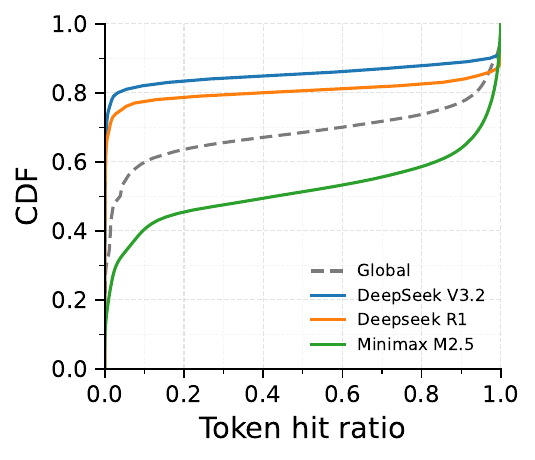}
        \caption{Model token hit ratio}
        \label{fig:model_cache_fraction_cdf}
    \end{subfigure}\hfill%
    \begin{subfigure}[t]{0.23\textwidth}
        \centering
        \includegraphics[width=\linewidth]{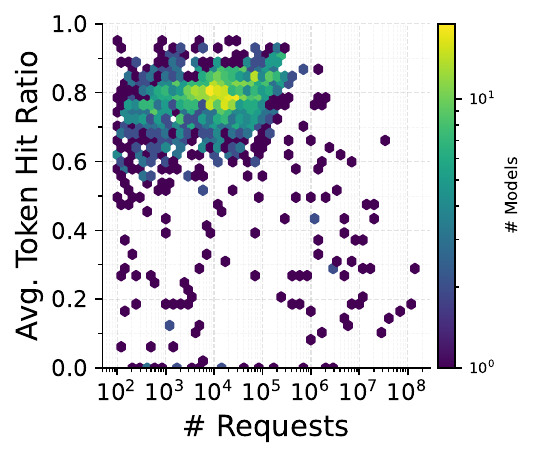}
        \caption{Per-model cache profile}
        \label{fig:model_cache_profiles}
    \end{subfigure}\hfill%
    \begin{subfigure}[t]{0.23\textwidth}
        \centering
        \includegraphics[width=\linewidth]{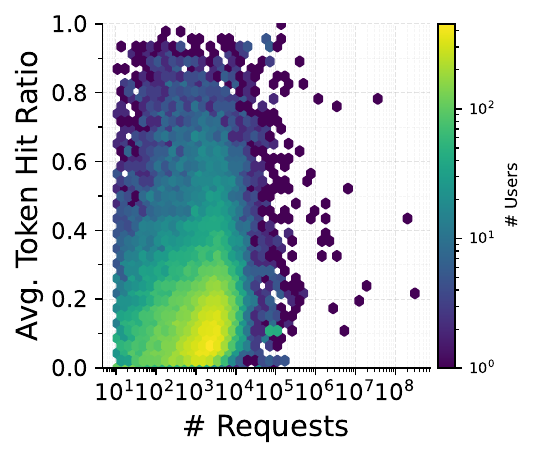}
        \caption{Per-user cache profile}
        \label{fig:user_cache_profiles}
    \end{subfigure}

    \vspace{-2mm}
    \caption{Cache behavior at three levels. Request-level cached-token fractions are bimodal and have distinctive characteristics across input lengths and model types. Models and users also have varying degrees of cacheability.}
    \vspace{-3mm}
    \label{fig:cache_behavior_overview}
\end{figure*}

\section{Prefix Caching}
\label{sec:caching}

Prefix caching reduces prefill work by reusing input tokens from earlier requests. Since prefill cost grows with input length, caching can substantially reduce TTFT. We first characterise reuse observed in production (\autoref{sec:caching_overview}) and the arrival locality that drives it (\autoref{sec:arrival_locality}); we then use simulation to compare eviction algorithms (\autoref{sec:caching_simulations}).

\subsection{Production System}
\label{sec:caching_overview}
We characterise production cache behavior using the logged cached-token field, which records the number of input tokens served from the prefix cache for each request. Thus, our analysis captures cache reuse actually realized by the deployed production system rather than theoretical. We analyze the final two months of the trace, during which cached-token logging was available. Figure~\ref{fig:cache_behavior_overview} presents the observed reuse from four perspectives: request-level distributions by input length and for selected high-volume models, together with cumulative per-model and per-user profiles.

\paragraph{Request-level cached fractions are bimodal.} Figure~\ref{fig:request_cache_fraction_cdf} shows a strongly bimodal pattern: most requests see either little to no prefix reuse or reuse nearly their entire input prefix, with relatively few requests falling in between. This pattern is consistent with many requests either reusing most of a prior prefix or receiving little reuse, as expected in multi-turn sessions. It shows that a request either inherits the previous turn's context (and therefore most of its prefix) or starts fresh, a partial hit being rare. The curves also shift with input size: short-input requests skew toward high reuse, typical of multi-turn dialogue where each turn appends an increment to a growing context, while long-input requests shift toward low reuse, since fewer of them inherit a previous turn that fully covers the new prefix.

\paragraph{Cache behavior differs sharply across models.} Figure~\ref{fig:model_cache_fraction_cdf} shows the request-level distribution for selected models. The two DeepSeek models have many requests with zero cache hits, while MiniMax-M2.5 shows a broader distribution, with many requests achieving intermediate to high hit ratios. Models with comparable traffic volume can therefore expose very different cacheability depending on how they are invoked. This may partly reflect MiniMax-M2.5's heavier use in coding or agentic workflows, which often involve short inter-arrival times and repeated reuse of prefixes across turns, leading to higher cache fractions.

\paragraph{Model and user cacheability.}
Figure~\ref{fig:model_cache_profiles} shows that several models with modest request volumes achieve high token hit ratios, suggesting that request volume alone does not determine reuse. Instead, these models may benefit from concentrated traffic: when a small set of users repeatedly accesses the same model with similar contexts, their requests can produce substantial reuse. Figure~\ref{fig:user_cache_profiles} supports this interpretation. Although most users exhibit little reuse, a smaller group achieves high hit ratios, likely by repeatedly accessing a limited set of models, low interarrival times, and proper context engineering. Model- and user-level cacheability are therefore intertwined. A model's aggregate hit ratio depends not only on the model itself, but also on its use case and user base. Models serving repetitive workloads or a concentrated set of users may achieve high reuse, whereas broadly used models with diverse requests may exhibit lower reuse. 

\subsection{Arrival Locality}
\label{sec:arrival_locality}

Caching is strongly influenced by \emph{arrival locality}: how quickly a user returns after a previous request. We first study recurrence at the user level, then refine it to the \emph{(user, model)} granularity, which better reflects cache reuse.

\begin{figure}[t]
    \centering

    \begin{subfigure}[t]{0.48\linewidth}
        \centering
        \includegraphics[width=\linewidth]{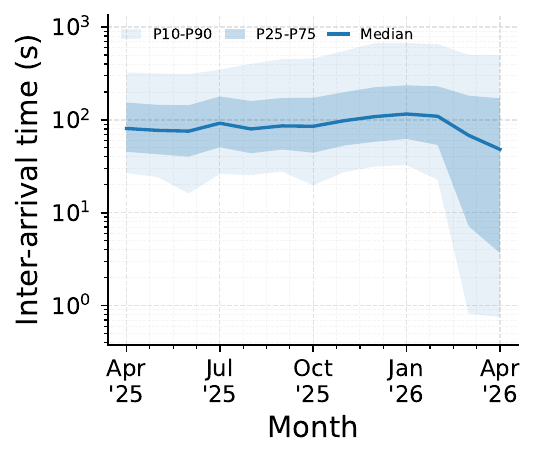}
        \caption{Per-user IAT over time}
        \label{fig:iat_per_user_monthly_trend}
    \end{subfigure}
    \hfill
    \begin{subfigure}[t]{0.48\linewidth}
        \centering
        \includegraphics[width=\linewidth]{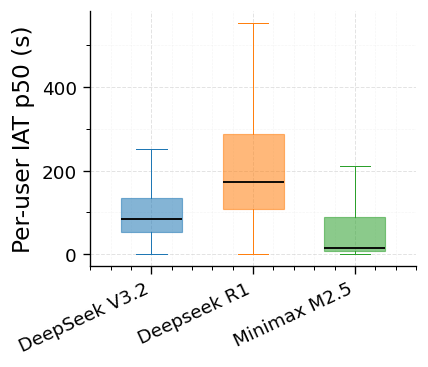}
        \caption{Per-user IAT across models}
        \label{fig:iat_per_user_by_model}
    \end{subfigure}
    \hfill
    \vspace{-2mm}
    \caption{Inter-arrival time (IAT) spread of users across different models. Per-user median trends downward over time.}
    \label{fig:iat_overview}
    \vspace{-4mm}
\end{figure}

\begin{figure*}[t]
    \centering
    \begin{minipage}[t]{0.328\linewidth}
        \centering
        \begin{subfigure}[t]{\linewidth}
        \includegraphics[width=\linewidth]{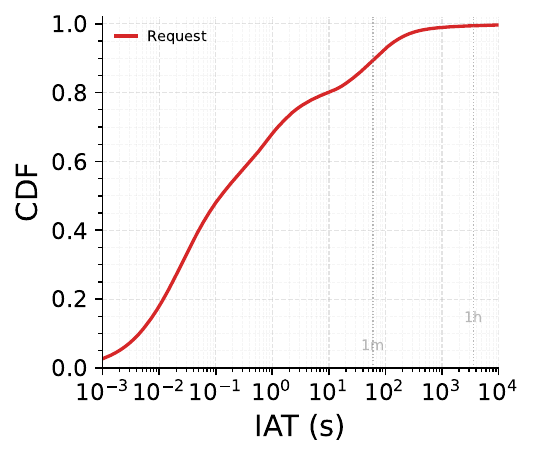}
        \caption{Per-user--model IAT}
        
        \end{subfigure}
        \captionof{figure}{99\% of reuse arrives within 15 minutes of the previous request. Cache locality is concentrated at short timescales.
        }
        \label{fig:pair_iat_cdf}
    \end{minipage}
    \hfill
    \begin{minipage}[t]{0.64\linewidth}
        \centering

        \begin{subfigure}[t]{0.48\linewidth}
            \centering
            \includegraphics[width=\linewidth]{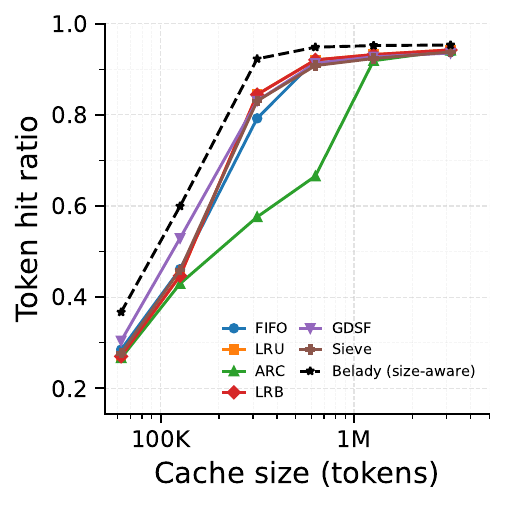}
            \caption{MiniMax-M2.5}
            \label{fig:minimax_m25_token_hit_ratio}
        \end{subfigure}
        \begin{subfigure}[t]{0.48\linewidth}
            \centering
            \includegraphics[width=\linewidth]{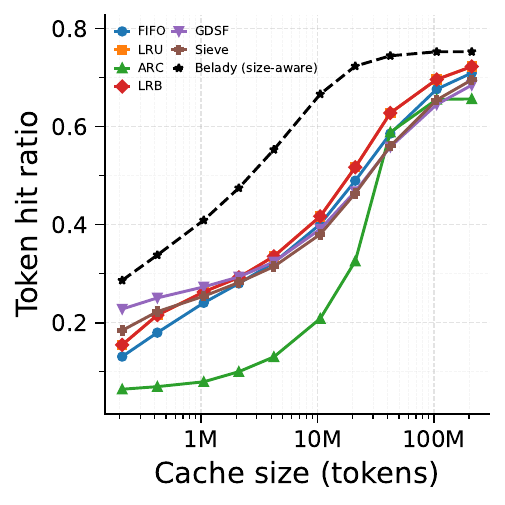}
            \caption{DeepSeek V3.2}
            \label{fig:deepseek_v32_token_hit_ratio}
        \end{subfigure}
        \hfill

        \vspace{-2mm}
        \captionof{figure}{
            Token hit ratio for seven eviction algorithms on workloads from two representative models. SOTA algorithms often perform similar or worse than FIFO and LRU. The size-aware Belady sets an upper bound for MiniMax-M2.5 but retains a persistent gap on DeepSeek V3.2.
        }
    \label{fig:others_token_hit_ratio}
    \end{minipage}
\end{figure*}

\paragraph{User IAT trends downwards over time.} Figure~\ref{fig:iat_per_user_monthly_trend} shows that per-user inter-arrival time remains on the order of minutes for most of the trace, but declines toward early 2026 with a widened spread. This indicates that users return more quickly over time, which might be partially explained with the growing adoption of more agents. The model-level view in Figure~\ref{fig:iat_per_user_by_model} shows that this recurrence pattern also varies substantially by model. MiniMax-M2.5 has the lowest per-user IAT, consistent with its use in coding or agentic workflows where requests are issued in rapid succession. DeepSeek~R1 has the longest IAT, which may reflect more human-paced reasoning interactions, where users wait for and read longer outputs before sending the next request. DeepSeek~V3.2 falls between these two extremes. Overall, user recurrence becomes faster over time, but the degree of arrival locality remains strongly workload-dependent.

\paragraph{A short TTL covers most repeat arrivals.}
Figure~\ref{fig:pair_iat_cdf} shows the inter-arrival time between consecutive requests from the same \emph{(user, model)} pair. Many requests return quickly with roughly half arrive within $0.1$ second, around 80\% within 10 seconds, and nearly all within several minutes. This strong temporal locality increases the likelihood that the previous turn's prefix remains in cache when the next request arrives. Thus, even a TTL of a few seconds can preserve reuse for a large share of repeat requests, while a TTL on the order of minutes captures nearly all request-weighted temporal locality.
\vspace{1em}
\begin{findingbox}
Cache reuse is common but uneven. Request-level cached fractions are bimodal and vary across models and users. Arrival locality is highly skewed but most repeat arrivals occurs at short timescales. 99\% requests return within a 15 minutes period.
\end{findingbox}
\subsection{Cache Simulations}
\label{sec:caching_simulations}

The analysis above measures the cached-token fraction in production systems, which are affected by both cache size and routing. 
In this section, we will try to evaluate the impact of different eviction algorithms by replaying the trace using cache simulations.

\subsubsection{Session Reconstruction}
\label{sec:caching_session_reconstruction}
Since the trace does not include prompts or explicit session identifiers, we reconstruct multi-turn sessions for simulation. Prior work~\cite{wang_kvcache_2025} shows that cross-user KV-cache hits are rare and that most reuse occurs within a user's own request stream. We therefore infer continuations only within each user.
For each request $B$ in arrival order, we keep a per-user sliding window containing the most recent $K{=}10$ earlier requests. We then select a unique unpaired candidate $A$ that satisfies three conditions: (1) $A$ uses the same API endpoint and model  as $B$; (2) $\mathit{it}_A + \mathit{ot}_A < \mathit{it}_B$, so the parent request's prompt and completion fit strictly inside the child request's input, as required by a valid multi-turn continuation; and (3) $A$ minimizes the residual gap $\mathit{it}_B - (\mathit{it}_A + \mathit{ot}_A)$, giving the tightest fit. If such a candidate exists, $B$ inherits $A$'s session ID and is assigned the next turn ID. Otherwise, $B$ starts a new session at turn 0.
This reconstruction deliberately ignores both the trace's recorded cached-token count and the production \texttt{instance\_id}. The inferred session structure therefore depends only on structural constraints implied by multi-turn conversations, not on the deployed router or cache policy. Consequently, the simulator estimates workload-intrinsic reuse rather than reuse induced by production routing locality.

\begin{figure*}[t]
    \centering
    \begin{subfigure}[t]{0.24\textwidth}
        \centering
        \includegraphics[width=\linewidth, trim=0 0 0 32, clip]{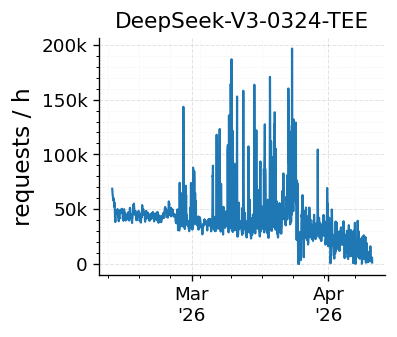}
        \caption{Requests / hour}
        \label{fig:case_lb_reqrate}
    \end{subfigure}
    \hfill
    \begin{subfigure}[t]{0.24\textwidth}
        \centering
        \includegraphics[width=\linewidth, trim=0 0 0 32, clip]{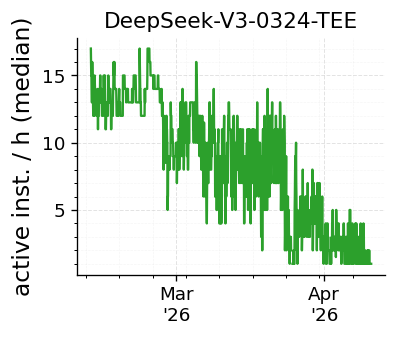}
        \caption{Active instances / hour}
        \label{fig:case_lb_active}
    \end{subfigure}
    \hfill
    \begin{subfigure}[t]{0.24\textwidth}
        \centering
        \includegraphics[width=\linewidth, trim=0 0 0 32, clip]{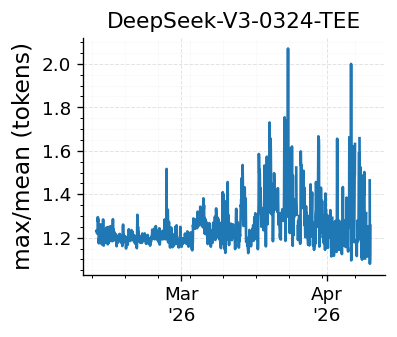}
        \caption{Token max/mean ratio}
        \label{fig:case_lb_maxmean}
    \end{subfigure}
    \hfill
    \begin{subfigure}[t]{0.24\textwidth}
        \centering
        \includegraphics[width=\linewidth, trim=0 0 0 32, clip]{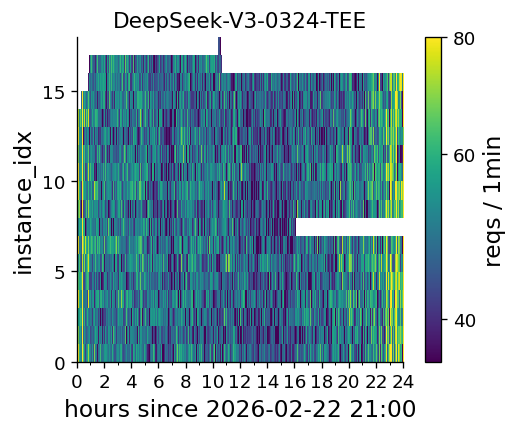}
        \caption{Per-instance occupancy}
        \label{fig:case_lb_concurrency}
    \end{subfigure}
    \vspace{-2mm}
    \caption{Workload and balance for DeepSeek-V3.2. (a) Request rate stays roughly stationary (b) while replicas trend downward. (c) Per-hour max/mean drifts upward as the fleet shrinks and (d) per-instance occupancy on the busiest day is balanced on average but episodically broken.}
    \vspace{-2mm}
    \label{fig:case_lb_demand_replication}
\end{figure*}

\subsubsection{Cache Simulation}
\label{sec:caching_single_cache_replay}
With sessions in hand, we first ask an upper-bound question: under a single cache, how much reuse is available in the workload, and how close do existing eviction algorithms come to the offline optimum? We replay the trace by treating each sequence as a management (admission and eviction) unit because \autoref{fig:request_cache_fraction_cdf} shows that most sequences either get zero or full hits.  
Using libCacheSim~\cite{libCacheSim}, we evaluate ten cache sizes per workload over several state-of-the-art algorithms: FIFO, LRU, ARC~\cite{ARC}, S3FIFO~\cite{s3fifo}, GDSF~\cite{cherkasova_improving_1998}, LHD~\cite{LHD}, LRU~\cite{LRB}, Sieve~\cite{sieve}, and the size-aware Belady oracle~\cite{belady_study_1966} algorithm. Due to limited space, we only present the ones with a larger difference. 
\vspace{2em}
\subsubsection{Results. }
Figure~\ref{fig:others_token_hit_ratio} highlights two results. First, simple eviction policies are already highly competitive for LLM prefix caching. LRU and FIFO often match or exceed the token hit ratios of more complex state-of-the-art algorithms. ARC is a notable exception: it performs substantially worse than LRU at intermediate cache sizes, especially on DeepSeek V3.2. This pattern is consistent with the strong recency bias in the workload as prefixes are frequently reused shortly after creation, but their utility decays quickly. 

Second, cache effectiveness is workload-dependent. DeepSeek V3.2 is more cache-constrained: its hit ratio increases gradually with cache size, and practical policies remain below the Belady oracle even at large capacities. MiniMax-M2.5 is more cache-friendly: its hit ratio increases quickly, and most policies approach the oracle at moderate cache sizes. This suggests that MiniMax-M2.5 has reusable session state that is easier to retain, whereas DeepSeek V3.2 has reuse patterns that are harder for standard eviction policies to capture.

\begin{findingbox}
Prefix-cache reuse is strongly recency-driven, making simple policies such as FIFO and LRU highly competitive. Nevertheless, the remaining gap to Belady reveals meaningful room for LLM-specific cache designs that better exploit session reuse and workload-dependent locality.
\end{findingbox}

\section{Load Balancing}

\subsection{Production Systems}
\label{sec:case_load_balancing}

\begin{figure}[t]
    \centering
    \begin{subfigure}[t]{0.49\linewidth}
        \centering
        \includegraphics[width=\linewidth, trim=0 0 0 32, clip]{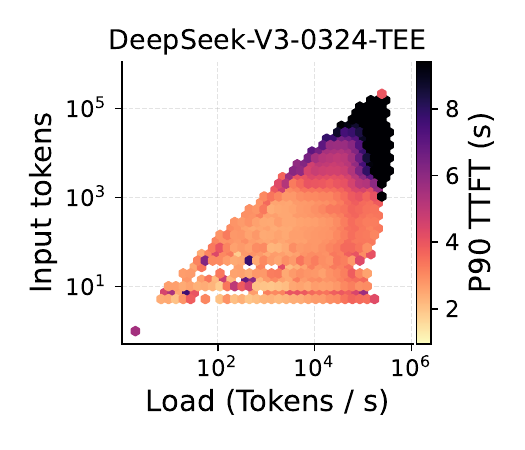}
        \caption{load vs TTFT}
        \label{fig:case_lb_ttft}
    \end{subfigure}
    \hfill
    \begin{subfigure}[t]{0.49\linewidth}
        \centering
        \includegraphics[width=\linewidth, trim=0 0 0 28, clip]{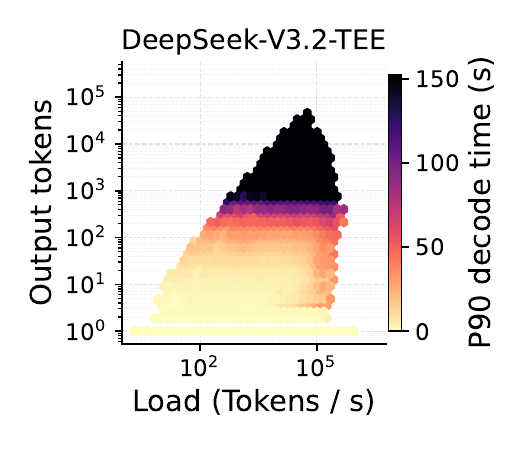}
        \caption{load vs decode}
        \label{fig:case_lb_duration}
    \end{subfigure}
    \vspace{-2mm}
    \caption{Per-request latency vs.\ instance-level load. TTFT and decode time climbs once long-token requests land on heavily loaded instances.}
    \vspace{-5mm}
    \label{fig:case_lb_latency}
\end{figure}

\begin{figure}[t]
    \centering
    \begin{subfigure}[t]{0.49\linewidth}
        \centering
        \includegraphics[width=\linewidth]{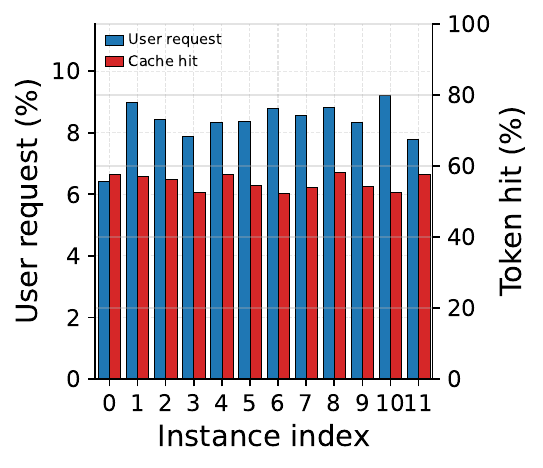}
        \caption{High load}
        \label{fig:case_cache_high_load}
    \end{subfigure}
    \hfill
    \begin{subfigure}[t]{0.49\linewidth}
        \centering
        \includegraphics[width=\linewidth]{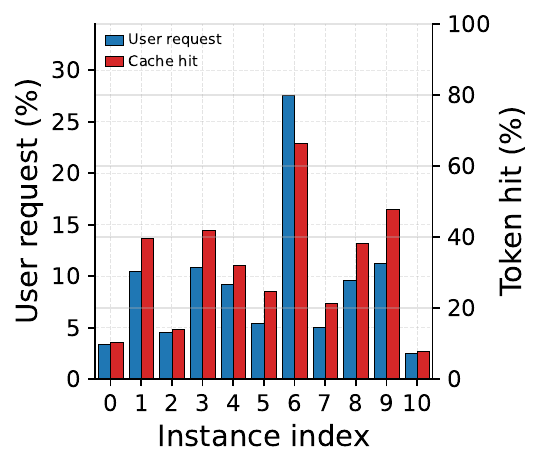}
        \caption{Low load}
        \label{fig:case_cache_low_load}
    \end{subfigure}
    \vspace{-2mm}
    \caption{Requests from a (user, model) pair over 60-minute windows. Under high load, routing spreads requests across instances, duplicating KV state; under low load, it concentrates them on fewer instances, preserving locality.}
    \label{fig:case_cache_regimes}
    \vspace{-6mm}
\end{figure}

We now ask whether requests within a single popular model are evenly distributed across the instances that serve it. We will be using DeepSeek-V3 as the case study throughout a 2-month period. 

\subsubsection{General Overview}
\label{sec:case_instance_balance}
Figure~\ref{fig:case_lb_reqrate} shows that request rate remains bursty but roughly stationary throughout the two months, while Figure~\ref{fig:case_lb_active} shows that the number of active instances declines. This decline closely follows the reduction in total token volume processed per hour, measured as input plus output tokens. Thus, replica count tracks serving utilization rather than request volume alone: scaling responds to token work, not just request arrivals.

Load is not always evenly distributed across active instances. Figure~\ref{fig:case_lb_maxmean} shows that the per-hour max/mean token-load ratio increases as the model scales down, indicating that the imbalance worsens as the serving pool shrinks. This imbalance matters because overload affects latency unevenly: Figure~\ref{fig:case_lb_ttft} shows that P90 TTFT rises sharply once an instance exceeds a high tokens-per-second threshold, especially for long-input requests. In contrast, Figure~\ref{fig:case_lb_duration} shows that P90 decode time mainly follows output length and is largely insensitive to instance throughput. Thus, overload primarily affects prefill, making input-heavy traffic the most vulnerable to load imbalance.

Zooming in on a single busy day reveals the short-term dynamics more clearly. Figure~\ref{fig:case_lb_concurrency} shows that most active instances remain busy and carry similar per-minute request rates, while a small number of instances become quiet around mid-day. Thus, load is balanced in aggregate, but not perfectly balanced at every instant.
This behavior reflects the production routing policy. For each request, the load balancer first selects the few least loaded instances, then routes to the one with the highest expected prefix-cache hits.

\subsubsection{Routing-Induced Cache Duplication}
\label{sec:case_routing_duplication}
Figure~\ref{fig:case_cache_regimes} shows how this tradeoff appears for the same user and model in two 60-minute windows. Under low load, Figure~\ref{fig:case_cache_low_load} shows that the user stays mostly on one instance, and cache hits concentrate there. Under high load, Figure~\ref{fig:case_cache_high_load} shows that the load balancer spreads the user across many instances to avoid overloading any one replica. The user’s traffic share then resembles the global load distribution, but several instances still show measurable hit ratios. 

These two regimes expose a fundamental tension between load balancing and cache locality. Keeping a user’s requests on one instance preserves its prefix state and maximizes reuse. Under high load, however, the router spreads requests across replicas. Cache hits may remain high, but only by replicating the same prefix across multiple instance-local KV caches. This duplication consumes scarce capacity and displaces prefixes from other users. A good routing algorithm must therefore balance traffic evenly while minimizing redundant KV state across the fleet.

\vspace{1em}
\begin{findingbox}
Stickiness preserves cache locality, but conflicts with load balancing. At high volume, the load balancer spreads a user across instances, duplicating the similar context across replica-local KV caches.
\end{findingbox}

\begin{figure*}[t]
    \centering

    \includegraphics[width=0.5\textwidth]{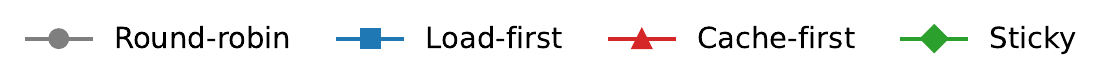}
    \vspace{-1mm}

    \begin{subfigure}[t]{0.235\textwidth}
        \centering
        \includegraphics[width=\linewidth]{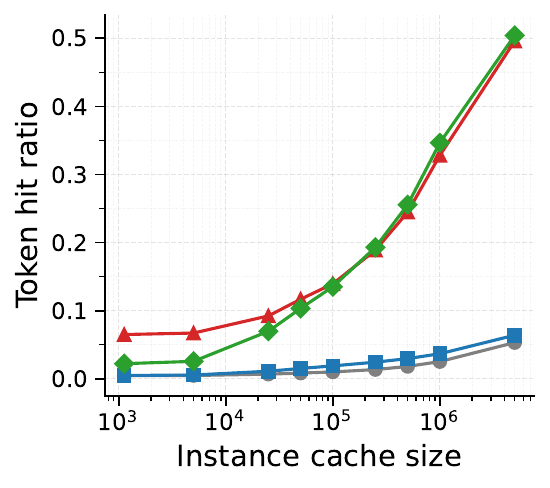}
    \end{subfigure}
    \hfill
    \begin{subfigure}[t]{0.235\textwidth}
        \centering
        \includegraphics[width=\linewidth]{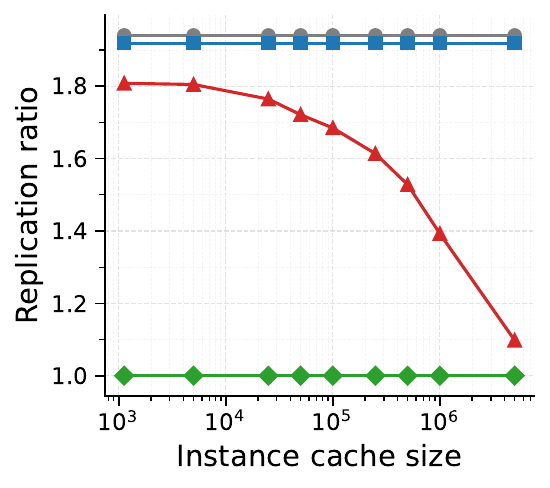}
    \end{subfigure}
    \hfill
    \begin{subfigure}[t]{0.235\textwidth}
        \centering
        \includegraphics[width=\linewidth]{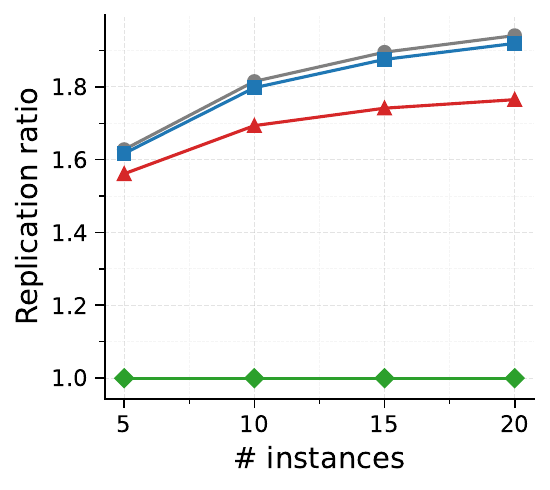}
    \end{subfigure}
    \hfill
    \begin{subfigure}[t]{0.235\textwidth}
        \centering
        \includegraphics[width=\linewidth]{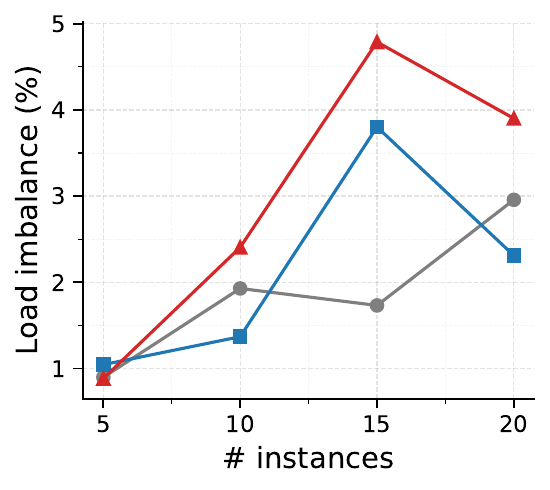}
    \end{subfigure}

    \vspace{-1mm}

    \begin{subfigure}[t]{0.235\textwidth}
        \centering
        \includegraphics[width=\linewidth]{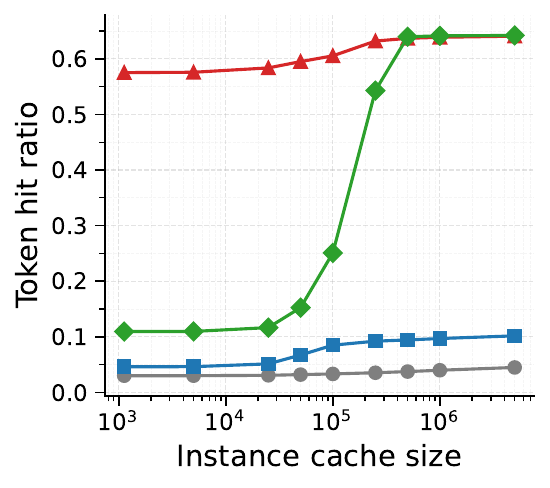}
        \caption{Hit ratio vs.\ cache size}
        \label{fig:sim_mm_hitrate_vs_cache}
    \end{subfigure}
    \hfill
    \begin{subfigure}[t]{0.235\textwidth}
        \centering
        \includegraphics[width=\linewidth]{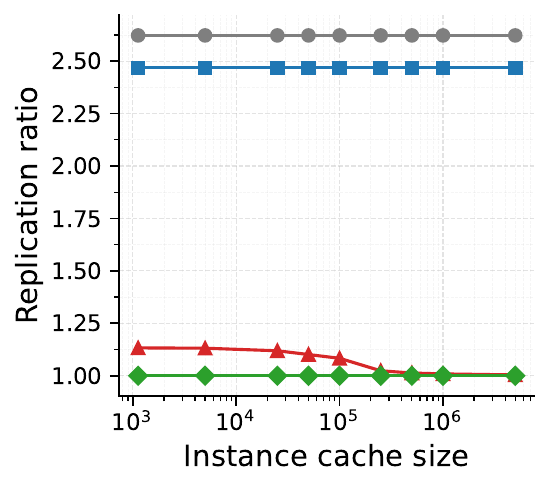}
        \caption{Replication vs.\ cache size}
        \label{fig:sim_mm_repl_vs_cache}
    \end{subfigure}
    \hfill
    \begin{subfigure}[t]{0.235\textwidth}
        \centering
        \includegraphics[width=\linewidth]{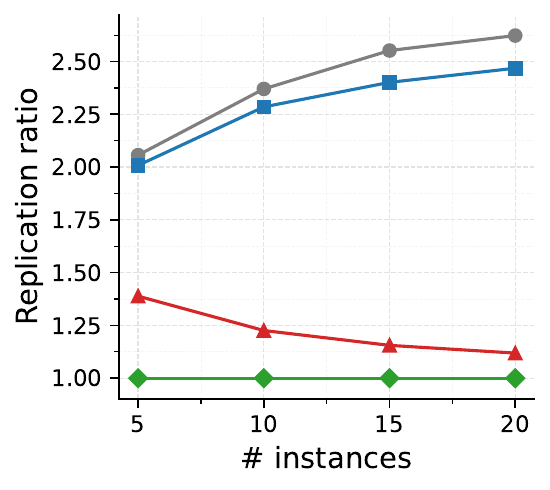}
        \caption{Replication vs.\ instance count}
        \label{fig:sim_mm_repl_vs_n}
    \end{subfigure}
    \hfill
    \begin{subfigure}[t]{0.235\textwidth}
        \centering
        \includegraphics[width=\linewidth]{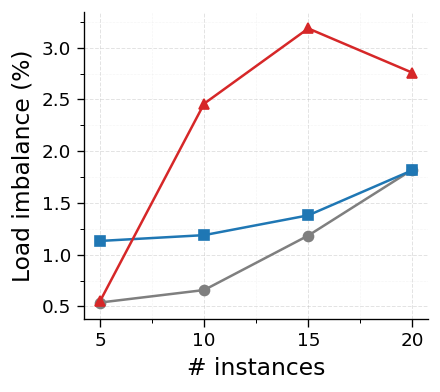}
        \caption{Imbalance vs.\ instance count}
        \label{fig:sim_mm_imbalance}
    \end{subfigure}

    \vspace{-2mm}
    \caption{Load balance simulation for DeepSeek V3.2 (top) and MiniMax-M2.5 (bottom). 
    (a) Cache-aware routing turns larger KV caches into higher hit ratios. 
    (b) Larger caches reduce replication for \textit{cache\_first}, while cache-blind policies remain highly replicated. 
    (c) Fleet scaling amplifies replication under cache-blind routing, but less than locality-aware routing. 
    (d) \textit{cache\_first} pays modest load imbalance (max/mean token/s); \textit{sticky} is omitted because its imbalance is orders of magnitude higher.}
    \label{fig:load_balancing_simulation_overview}
    \vspace{-2mm}
\end{figure*}

\subsection{Load Balancing Simulation}
\label{sec:case_routing_simulation}
Our production analysis illustrates the tension between cache locality and load balance. Under low load, a heavy user remains on one instance and benefits from repeated cache hits. Under high load, the load balancer spreads the user's requests across the fleet, duplicating KV state and reducing the maximum achievable cache reuse. Prior work addresses this tension through \emph{KV-cache transfer and offloading}~\cite{qin_mooncake_2025,liu2025lmcache}, which stores KV state outside the serving instance or transfers it between workers as routing changes.

Although these designs preserve reuse across instances, they turn cache locality into a data-movement problem. KV state can be very large for long-context requests. For example, a 100K-token MiniMax-M2.5 context occupies roughly 27 GB of KV state, requiring about 2.7 s to transfer over a 10 GB/s link in one direction, excluding contention, coordination, and any additional transfer to the destination worker. At the same time, our trace shows that reuse often occurs over short timescales, with many repeat requests from the same \emph{(user, model)} pair arriving within seconds. In such cases, KV transfer may impose substantial network overhead for limited latency benefit.

We therefore study a complementary approach: preserve locality by routing requests to instances that already hold useful KV state, and relax this preference only when load imbalance becomes too large. This avoids explicit KV transfer on the critical path and keeps reuse in local GPU memory, where it directly reduces repeated prefill work. The central scheduling question is therefore how much load imbalance should be tolerated to preserve cache locality. To quantify this tradeoff, we replay the chained trace from \autoref{sec:caching_simulations} on a simulated fleet of $N$ instances, each with a local LRU KV cache. We ask two questions: how much cache hit ratio does balance-oriented routing sacrifice, and how much load imbalance does locality-oriented routing introduce?

\subsubsection{Simulator and Routing Policies}
\label{sec:case_routing_setup} 
The simulator processes requests in arrival order, but now over a fleet of instances that acts as a distributed cache. Each instance manages its own KV cache and in-flight load independently. Before routing each request, the simulator advances time to the request arrival timestamp and removes completed requests from the active-load counters. It then applies the routing policy, performs a session-level cache lookup on the chosen instance, records the reusable prefix tokens, adds the request to that instance's in-flight queue until completion, and updates its local KV cache with the new session footprint. This produces a time-aware replay in which routing decisions, cache state, and per-instance are recorded.

Four routing policies span the locality--balance spectrum. \textit{round\_\allowbreak robin} rotates requests modulo $N$ and ignores both load and cache. \textit{load\_first} sends each request to the instance with the smallest current routing load, where load is measured by total tokens. \textit{sticky} pins each user to the first instance it reaches, approximating naive locality without prefix awareness. \textit{cache\_first} chooses the instance with the largest reusable prefix for the request's session, falling back to the least-loaded instance when no cached state exists.

We reuse the reconstructed multi-turn sessions from \autoref{sec:caching_simulations}. Based on the single-cache simulation in \autoref{fig:others_token_hit_ratio}, we use LRU for the routing study. Since the eviction policies perform similarly, LRU provides a simple and representative cache baseline. We also select the same two models used previously, with DeepSeek V3.2 having a larger working set that sustains cache pressure across the values that are swept. MiniMax-M2.5 is much more cacheable, and its sessions fit more comfortably in replica-local KV caches. Together these two models therefore expose both a cache-constrained workload and a more cache-friendly workload. We sweep per-instance cache size in $[1.1\text{K}, 5\text{M}]$ tokens and fleet size $N \in \{5, 10, 15, 20\}$.

\subsubsection{Simulation Results}
\label{sec:case_routing_results}
Figure~\ref{fig:load_balancing_simulation_overview} traces the routing tradeoff from cache reuse, to KV duplication, to load balance. We first ask whether a larger per-instance KV cache actually improves reuse under each routing policy. We then ask where that reused state lives: concentrated on one replica, or duplicated across the fleet. Finally, we measure the load-balance cost of preserving locality. 

\paragraph{\textbf{Cache-aware routing unlocks fleet-wide KV capacity.}}
Figure~\ref{fig:load_balancing_simulation_overview}(a) shows the token hit ratio as per-instance cache size increases at $N{=}20$. The main distinction is between cache-aware and cache-blind routing. \textit{load\_first} and \textit{round\_robin} maintain low hit ratios even as cache capacity grows because they scatter future turns across replicas; additional capacity provides little benefit when requests do not return to the instance holding their prefix. In contrast, \textit{cache\_first} converts larger caches into substantially higher reuse and approaches the single-cache headroom in \autoref{fig:others_token_hit_ratio}. \textit{sticky} also preserves locality, but only at user granularity. Because each user is pinned to one instance, a heavy user can access only that replica's cache rather than the fleet's aggregate capacity. This effect is most visible for MiniMax-M2.5: \textit{cache\_first} achieves high hit ratios even with small per-instance caches, whereas \textit{sticky} improves only once each replica can hold the working sets of its pinned users. For DeepSeek V3.2, the larger working set causes both locality-aware policies to continue improving as cache size grows. 

\paragraph{\textbf{Larger caches strengthen cache affinity and reduce session spread.}}
Figure~\ref{fig:load_balancing_simulation_overview}(b) partially explains the hit-ratio gap at $N{=}20$. We define the replication ratio of a session as the number of distinct instances that serve its requests; a value of 1.0 means that the entire session remains on one instance. Higher values indicate that the session is spread across more replicas, increasing the likelihood that its KV state is recomputed or stored at multiple locations. \textit{sticky} remains at 1.0 by construction because each user is pinned to one instance. In contrast, \textit{load\_first} and \textit{round\_robin} maintain high replication ratios across cache sizes because their routing decisions ignore cached state. For \textit{cache\_first}, replication is high when caches are small. Prefixes are frequently evicted before later turns arrive, leaving no cached instance to preserve affinity and forcing the policy to fall back to least-loaded routing. Successive turns may therefore visit different instances. As cache capacity grows, prefixes remain resident longer, later turns return to the same instance, and the replication ratio approaches 1.0. This transition occurs later for DeepSeek V3.2, whose larger working set requires more capacity, and earlier for MiniMax-M2.5, whose reusable state is easier to retain. Thus, larger caches allow \textit{cache\_first} to improve reuse while keeping each session concentrated on fewer replicas.

\paragraph{\textbf{Scaling amplifies the session replication cost of cache-blind routing.}} Figure~\ref{fig:load_balancing_simulation_overview}(c) fixes the per-instance cache size at 25k and varies the number of serving instances. \textit{sticky} remains at 1.0 by construction because each user is pinned to one replica. In contrast, the replication ratios of \textit{round\_robin} and \textit{load\_first} increase steadily with $N$ on both workloads. As the fleet grows, successive turns from the same session can be routed to more instances, spreading the session across more replica-local caches. The effect of scaling on \textit{cache\_first} is workload-dependent. For DeepSeek V3.2, replication increases moderately with fleet size because its larger working set remains difficult to retain at the fixed cache size. When a prefix is evicted, later turns lose their cache-affinity signal and fall back to least-loaded routing, causing the session to visit additional instances. For MiniMax-M2.5, replication instead decreases from roughly 1.4 toward 1.1. Its smaller reusable working set fits more easily as the fleet grows and the workload is divided across more caches, allowing prefixes to remain resident and later turns to return to the same instance. Thus, scaling reduces session replication under cache-aware routing only when the distributed cache capacity is sufficient to preserve useful prefixes.

\paragraph{\textbf{Locality has a bounded load-balancing cost.}}
Figure~\ref{fig:load_balancing_simulation_overview}(d) measures the imbalance introduced by preserving cache locality. We omit \textit{sticky}, since pinning entire users to fixed instances produces a far larger imbalance and would obscure the other policies. Among the remaining policies, \textit{round\_robin} and \textit{load\_first} are generally the most balanced, while \textit{cache\_first} incurs higher imbalance because it prefers replicas that already hold reusable session state. This is the price of locality: some requests are intentionally routed away from the least loaded instance to avoid losing cache reuse. 
The absolute imbalance is surprisingly small, remaining within 5\% to 7\%. We find that this is largely because the workload contains many single-turn sessions, which \textit{cache\_first} can freely route to the least-loaded instance to offset localized multi-turn traffic. Workloads with fewer one-off requests may provide less balancing flexibility. Future routing policies should therefore preserve session locality when possible, while using load-aware fallback or session migration to control imbalance without naively scattering KV state across replicas.

\vspace{1em}
\begin{findingbox}
LLM serving faces a fundamental tension between cache locality and load balance. Cache-aware routing improves reuse and reduces cross-instance KV duplication with modest imbalance. Larger caches alone cannot recover reuse when routing scatters session state across replicas.
\end{findingbox}

\section{Related Work}

\paragraph{LLM Inference Optimization.}
A large body of work improves LLM serving through better batching, scheduling, memory management, prefill--decode separation, and KV-cache reuse. 
Orca, vLLM, and Sarathi-Serve improve continuous batching, KV-cache memory management, and prefill scheduling~\cite{yu_orca_2022,kwon_vllm_2023,agrawal_sarathi_2024}. 
DistServe, Splitwise, and Mooncake disaggregate prefill, decode, or KV-cache management to improve serving efficiency~\cite{zhong_distserve_2024,patel_splitwise_2024,qin_mooncake_2025}. 
Other work targets cache reuse more directly: H$_2$O proposes an eviction policy, Prompt Cache and CacheBlend reuse cached computation, SGLang introduces RadixAttention for prefix reuse, and Preble jointly considers cache locality and distributed load balancing~\cite{zhang_h2o_2023,gim_promptcache_2024,yao_cacheblend_2025,zheng_sglang_2024,srivatsa_preble_2025}.

These optimizations depend on realistic datasets for faithful evaluation. Batching and scheduling depend on arrival patterns and token lengths; disaggregated serving depends on the balance between prefill and decode; and cache-aware routing depends on temporal locality and prefix reuse. Yet evaluations often rely on benchmark or application-specific workloads, such as ShareGPT-derived requests, chat, code-completion, and summarization traces, or short and partially released production traces~\cite{kwon_vllm_2023,zheng_sglang_2024,zhong_distserve_2024,patel_splitwise_2024,srivatsa_preble_2025}. These workloads often require synthetic arrivals or lose timing fidelity through sampling, obscuring inter-arrival times, and losing user-level information. Our one-year, request-level production trace preserves these signals enabling more realistic evaluation of serving-system designs.

\section{Conclusion}

We present and release a one-year production trace of LLM serving and use it to characterize workload evolution, user--model interactions, prefix caching, and load-balancing behavior. Our analysis shows that production LLM workloads are dynamic and heterogeneous: demand evolves over time, users and models exhibit diverse usage patterns, and cache locality is unevenly distributed across the workload. We further find that prefix caching and load balancing are closely coupled, creating a fundamental tradeoff between preserving reuse and distributing requests evenly. These results highlight the need for trace-driven serving-system designs that account for temporal evolution, workload structure, and cache-aware routing. 

\newpage

\bibliographystyle{ACM-Reference-Format}

\begin{thebibliography}{38}


\ifx \showCODEN    \undefined \def \showCODEN     #1{\unskip}     \fi
\ifx \showISBNx    \undefined \def \showISBNx     #1{\unskip}     \fi
\ifx \showISBNxiii \undefined \def \showISBNxiii  #1{\unskip}     \fi
\ifx \showISSN     \undefined \def \showISSN      #1{\unskip}     \fi
\ifx \showLCCN     \undefined \def \showLCCN      #1{\unskip}     \fi
\ifx \shownote     \undefined \def \shownote      #1{#1}          \fi
\ifx \showarticletitle \undefined \def \showarticletitle #1{#1}   \fi
\ifx \showURL      \undefined \def \showURL       {\relax}        \fi
\providecommand\bibfield[2]{#2}
\providecommand\bibinfo[2]{#2}
\providecommand\natexlab[1]{#1}
\providecommand\showeprint[2][]{arXiv:#2}

\bibitem[lib({[n.\,d.]})]%
        {libCacheSim}
 \bibinfo{year}{[n.\,d.]}\natexlab{}.
\newblock \bibinfo{title}{libCacheSim: A high-performance cache simulator}.
\newblock \bibinfo{howpublished}{\url{https://github.com/1a1a11a/libCacheSim}}.
\newblock
\newblock
\shownote{Accessed: 2025-12-01}.


\bibitem[Agrawal et~al\mbox{.}(2024)]%
        {agrawal_sarathi_2024}
\bibfield{author}{\bibinfo{person}{Amey Agrawal}, \bibinfo{person}{Nitin
  Kedia}, \bibinfo{person}{Ashish Panwar}, \bibinfo{person}{Jayashree Mohan},
  \bibinfo{person}{Nipun Kwatra}, \bibinfo{person}{Bhargav~S. Gulavani},
  \bibinfo{person}{Alexey Tumanov}, {and} \bibinfo{person}{Ramachandran
  Ramjee}.} \bibinfo{year}{2024}\natexlab{}.
\newblock \showarticletitle{Taming Throughput-Latency Tradeoff in LLM Inference
  with Sarathi-Serve}. In \bibinfo{booktitle}{\emph{18th USENIX Symposium on
  Operating Systems Design and Implementation (OSDI 24)}}.
  \bibinfo{pages}{117--134}.
\newblock


\bibitem[Beckmann et~al\mbox{.}(2018)]%
        {LHD}
\bibfield{author}{\bibinfo{person}{Nathan Beckmann}, \bibinfo{person}{Haoxian
  Chen}, {and} \bibinfo{person}{Asaf Cidon}.} \bibinfo{year}{2018}\natexlab{}.
\newblock \showarticletitle{{LHD:} Improving Cache Hit Rate by Maximizing Hit
  Density}. In \bibinfo{booktitle}{\emph{15th {USENIX} Symposium on Networked
  Systems Design and Implementation, {NSDI} 2018, Renton, WA, USA, April 9-11,
  2018}}, \bibfield{editor}{\bibinfo{person}{Sujata Banerjee} {and}
  \bibinfo{person}{Srinivasan Seshan}} (Eds.). \bibinfo{publisher}{{USENIX}
  Association}, \bibinfo{pages}{389--403}.
\newblock
\urldef\tempurl%
\url{https://www.usenix.org/conference/nsdi18/presentation/beckmann}
\showURL{%
\tempurl}


\bibitem[Belady(1966)]%
        {belady_study_1966}
\bibfield{author}{\bibinfo{person}{L.~A. Belady}.}
  \bibinfo{year}{1966}\natexlab{}.
\newblock \showarticletitle{A study of replacement algorithms for a
  virtual-storage computer}.
\newblock \bibinfo{journal}{\emph{IBM Systems Journal}} \bibinfo{volume}{5},
  \bibinfo{number}{2} (\bibinfo{year}{1966}), \bibinfo{pages}{78--101}.
\newblock
\showISSN{0018-8670}
\href{https://doi.org/10.1147/sj.52.0078}{doi:\nolinkurl{10.1147/sj.52.0078}}


\bibitem[Cherkasova(1998)]%
        {cherkasova_improving_1998}
\bibfield{author}{\bibinfo{person}{Ludmila Cherkasova}.}
  \bibinfo{year}{1998}\natexlab{}.
\newblock \bibinfo{booktitle}{\emph{Improving {WWW} proxies performance with
  greedy-dual-size-frequency caching policy}}.
\newblock \bibinfo{publisher}{Citeseer}.
\newblock


\bibitem[Chung et~al\mbox{.}(2020)]%
        {chung2020unearthing}
\bibfield{author}{\bibinfo{person}{Andrew Chung}, \bibinfo{person}{Subru
  Krishnan}, \bibinfo{person}{Konstantinos Karanasos}, \bibinfo{person}{Carlo
  Curino}, {and} \bibinfo{person}{Gregory~R Ganger}.}
  \bibinfo{year}{2020}\natexlab{}.
\newblock \showarticletitle{Unearthing inter-job dependencies for better
  cluster scheduling}. In \bibinfo{booktitle}{\emph{14th USENIX Symposium on
  Operating Systems Design and Implementation (OSDI 20)}}.
  \bibinfo{pages}{1205--1223}.
\newblock


\bibitem[DeepSeek-AI et~al\mbox{.}(2025a)]%
        {deepseek-ai_deepseek-r1_2025}
\bibfield{author}{\bibinfo{person}{DeepSeek-AI}, \bibinfo{person}{Daya Guo},
  \bibinfo{person}{Dejian Yang}, \bibinfo{person}{Haowei Zhang},
  \bibinfo{person}{Junxiao Song}, \bibinfo{person}{Ruoyu Zhang},
  \bibinfo{person}{Runxin Xu}, \bibinfo{person}{Qihao Zhu},
  \bibinfo{person}{Shirong Ma}, \bibinfo{person}{Peiyi Wang},
  \bibinfo{person}{Xiao Bi}, \bibinfo{person}{Xiaokang Zhang},
  \bibinfo{person}{Xingkai Yu}, \bibinfo{person}{Yu Wu}, \bibinfo{person}{Z.~F.
  Wu}, \bibinfo{person}{Zhibin Gou}, \bibinfo{person}{Zhihong Shao},
  \bibinfo{person}{Zhuoshu Li}, \bibinfo{person}{Ziyi Gao},
  \bibinfo{person}{Aixin Liu}, \bibinfo{person}{Bing Xue},
  \bibinfo{person}{Bingxuan Wang}, \bibinfo{person}{Bochao Wu},
  \bibinfo{person}{Bei Feng}, \bibinfo{person}{Chengda Lu},
  \bibinfo{person}{Chenggang Zhao}, \bibinfo{person}{Chengqi Deng},
  \bibinfo{person}{Chenyu Zhang}, \bibinfo{person}{Chong Ruan},
  \bibinfo{person}{Damai Dai}, \bibinfo{person}{Deli Chen},
  \bibinfo{person}{Dongjie Ji}, \bibinfo{person}{Erhang Li},
  \bibinfo{person}{Fangyun Lin}, \bibinfo{person}{Fucong Dai},
  \bibinfo{person}{Fuli Luo}, \bibinfo{person}{Guangbo Hao},
  \bibinfo{person}{Guanting Chen}, \bibinfo{person}{Guowei Li},
  \bibinfo{person}{H. Zhang}, \bibinfo{person}{Han Bao},
  \bibinfo{person}{Hanwei Xu}, \bibinfo{person}{Haocheng Wang},
  \bibinfo{person}{Honghui Ding}, \bibinfo{person}{Huajian Xin},
  \bibinfo{person}{Huazuo Gao}, \bibinfo{person}{Hui Qu}, \bibinfo{person}{Hui
  Li}, \bibinfo{person}{Jianzhong Guo}, \bibinfo{person}{Jiashi Li},
  \bibinfo{person}{Jiawei Wang}, \bibinfo{person}{Jingchang Chen},
  \bibinfo{person}{Jingyang Yuan}, \bibinfo{person}{Junjie Qiu},
  \bibinfo{person}{Junlong Li}, \bibinfo{person}{J.~L. Cai},
  \bibinfo{person}{Jiaqi Ni}, \bibinfo{person}{Jian Liang},
  \bibinfo{person}{Jin Chen}, \bibinfo{person}{Kai Dong}, \bibinfo{person}{Kai
  Hu}, \bibinfo{person}{Kaige Gao}, \bibinfo{person}{Kang Guan},
  \bibinfo{person}{Kexin Huang}, \bibinfo{person}{Kuai Yu},
  \bibinfo{person}{Lean Wang}, \bibinfo{person}{Lecong Zhang},
  \bibinfo{person}{Liang Zhao}, \bibinfo{person}{Litong Wang},
  \bibinfo{person}{Liyue Zhang}, \bibinfo{person}{Lei Xu},
  \bibinfo{person}{Leyi Xia}, \bibinfo{person}{Mingchuan Zhang},
  \bibinfo{person}{Minghua Zhang}, \bibinfo{person}{Minghui Tang},
  \bibinfo{person}{Meng Li}, \bibinfo{person}{Miaojun Wang},
  \bibinfo{person}{Mingming Li}, \bibinfo{person}{Ning Tian},
  \bibinfo{person}{Panpan Huang}, \bibinfo{person}{Peng Zhang},
  \bibinfo{person}{Qiancheng Wang}, \bibinfo{person}{Qinyu Chen},
  \bibinfo{person}{Qiushi Du}, \bibinfo{person}{Ruiqi Ge},
  \bibinfo{person}{Ruisong Zhang}, \bibinfo{person}{Ruizhe Pan},
  \bibinfo{person}{Runji Wang}, \bibinfo{person}{R.~J. Chen},
  \bibinfo{person}{R.~L. Jin}, \bibinfo{person}{Ruyi Chen},
  \bibinfo{person}{Shanghao Lu}, \bibinfo{person}{Shangyan Zhou},
  \bibinfo{person}{Shanhuang Chen}, \bibinfo{person}{Shengfeng Ye},
  \bibinfo{person}{Shiyu Wang}, \bibinfo{person}{Shuiping Yu},
  \bibinfo{person}{Shunfeng Zhou}, \bibinfo{person}{Shuting Pan},
  \bibinfo{person}{S.~S. Li}, \bibinfo{person}{Shuang Zhou},
  \bibinfo{person}{Shaoqing Wu}, \bibinfo{person}{Shengfeng Ye},
  \bibinfo{person}{Tao Yun}, \bibinfo{person}{Tian Pei},
  \bibinfo{person}{Tianyu Sun}, \bibinfo{person}{T. Wang},
  \bibinfo{person}{Wangding Zeng}, \bibinfo{person}{Wanjia Zhao},
  \bibinfo{person}{Wen Liu}, \bibinfo{person}{Wenfeng Liang},
  \bibinfo{person}{Wenjun Gao}, \bibinfo{person}{Wenqin Yu},
  \bibinfo{person}{Wentao Zhang}, \bibinfo{person}{W.~L. Xiao},
  \bibinfo{person}{Wei An}, \bibinfo{person}{Xiaodong Liu},
  \bibinfo{person}{Xiaohan Wang}, \bibinfo{person}{Xiaokang Chen},
  \bibinfo{person}{Xiaotao Nie}, \bibinfo{person}{Xin Cheng},
  \bibinfo{person}{Xin Liu}, \bibinfo{person}{Xin Xie},
  \bibinfo{person}{Xingchao Liu}, \bibinfo{person}{Xinyu Yang},
  \bibinfo{person}{Xinyuan Li}, \bibinfo{person}{Xuecheng Su},
  \bibinfo{person}{Xuheng Lin}, \bibinfo{person}{X.~Q. Li},
  \bibinfo{person}{Xiangyue Jin}, \bibinfo{person}{Xiaojin Shen},
  \bibinfo{person}{Xiaosha Chen}, \bibinfo{person}{Xiaowen Sun},
  \bibinfo{person}{Xiaoxiang Wang}, \bibinfo{person}{Xinnan Song},
  \bibinfo{person}{Xinyi Zhou}, \bibinfo{person}{Xianzu Wang},
  \bibinfo{person}{Xinxia Shan}, \bibinfo{person}{Y.~K. Li},
  \bibinfo{person}{Y.~Q. Wang}, \bibinfo{person}{Y.~X. Wei},
  \bibinfo{person}{Yang Zhang}, \bibinfo{person}{Yanhong Xu},
  \bibinfo{person}{Yao Li}, \bibinfo{person}{Yao Zhao},
  \bibinfo{person}{Yaofeng Sun}, \bibinfo{person}{Yaohui Wang},
  \bibinfo{person}{Yi Yu}, \bibinfo{person}{Yichao Zhang},
  \bibinfo{person}{Yifan Shi}, \bibinfo{person}{Yiliang Xiong},
  \bibinfo{person}{Ying He}, \bibinfo{person}{Yishi Piao},
  \bibinfo{person}{Yisong Wang}, \bibinfo{person}{Yixuan Tan},
  \bibinfo{person}{Yiyang Ma}, \bibinfo{person}{Yiyuan Liu},
  \bibinfo{person}{Yongqiang Guo}, \bibinfo{person}{Yuan Ou},
  \bibinfo{person}{Yuduan Wang}, \bibinfo{person}{Yue Gong},
  \bibinfo{person}{Yuheng Zou}, \bibinfo{person}{Yujia He},
  \bibinfo{person}{Yunfan Xiong}, \bibinfo{person}{Yuxiang Luo},
  \bibinfo{person}{Yuxiang You}, \bibinfo{person}{Yuxuan Liu},
  \bibinfo{person}{Yuyang Zhou}, \bibinfo{person}{Y.~X. Zhu},
  \bibinfo{person}{Yanhong Xu}, \bibinfo{person}{Yanping Huang},
  \bibinfo{person}{Yaohui Li}, \bibinfo{person}{Yi Zheng},
  \bibinfo{person}{Yuchen Zhu}, \bibinfo{person}{Yunxian Ma},
  \bibinfo{person}{Ying Tang}, \bibinfo{person}{Yukun Zha},
  \bibinfo{person}{Yuting Yan}, \bibinfo{person}{Z.~Z. Ren},
  \bibinfo{person}{Zehui Ren}, \bibinfo{person}{Zhangli Sha},
  \bibinfo{person}{Zhe Fu}, \bibinfo{person}{Zhean Xu}, \bibinfo{person}{Zhenda
  Xie}, \bibinfo{person}{Zhengyan Zhang}, \bibinfo{person}{Zhewen Hao},
  \bibinfo{person}{Zhicheng Ma}, \bibinfo{person}{Zhigang Yan},
  \bibinfo{person}{Zhiyu Wu}, \bibinfo{person}{Zihui Gu},
  \bibinfo{person}{Zijia Zhu}, \bibinfo{person}{Zijun Liu},
  \bibinfo{person}{Zilin Li}, \bibinfo{person}{Ziwei Xie},
  \bibinfo{person}{Ziyang Song}, \bibinfo{person}{Zizheng Pan},
  \bibinfo{person}{Zhen Huang}, \bibinfo{person}{Zhipeng Xu},
  \bibinfo{person}{Zhongyu Zhang}, {and} \bibinfo{person}{Zhen Zhang}.}
  \bibinfo{year}{2025}\natexlab{a}.
\newblock \bibinfo{title}{{DeepSeek}-{R1}: {Incentivizing} {Reasoning}
  {Capability} in {LLMs} via {Reinforcement} {Learning}}.
\newblock
\href{https://doi.org/10.48550/arXiv.2501.12948}{doi:\nolinkurl{10.48550/arXiv.2501.12948}}
\newblock
\shownote{arXiv:2501.12948 [cs]}.


\bibitem[DeepSeek-AI et~al\mbox{.}(2025b)]%
        {deepseek-ai_deepseek-v3_2025}
\bibfield{author}{\bibinfo{person}{DeepSeek-AI}, \bibinfo{person}{Aixin Liu},
  \bibinfo{person}{Bei Feng}, \bibinfo{person}{Bing Xue},
  \bibinfo{person}{Bingxuan Wang}, \bibinfo{person}{Bochao Wu},
  \bibinfo{person}{Chengda Lu}, \bibinfo{person}{Chenggang Zhao},
  \bibinfo{person}{Chengqi Deng}, \bibinfo{person}{Chenyu Zhang},
  \bibinfo{person}{Chong Ruan}, \bibinfo{person}{Damai Dai},
  \bibinfo{person}{Daya Guo}, \bibinfo{person}{Dejian Yang},
  \bibinfo{person}{Deli Chen}, \bibinfo{person}{Dongjie Ji},
  \bibinfo{person}{Erhang Li}, \bibinfo{person}{Fangyun Lin},
  \bibinfo{person}{Fucong Dai}, \bibinfo{person}{Fuli Luo},
  \bibinfo{person}{Guangbo Hao}, \bibinfo{person}{Guanting Chen},
  \bibinfo{person}{Guowei Li}, \bibinfo{person}{H. Zhang}, \bibinfo{person}{Han
  Bao}, \bibinfo{person}{Hanwei Xu}, \bibinfo{person}{Haocheng Wang},
  \bibinfo{person}{Haowei Zhang}, \bibinfo{person}{Honghui Ding},
  \bibinfo{person}{Huajian Xin}, \bibinfo{person}{Huazuo Gao},
  \bibinfo{person}{Hui Li}, \bibinfo{person}{Hui Qu}, \bibinfo{person}{J.~L.
  Cai}, \bibinfo{person}{Jian Liang}, \bibinfo{person}{Jianzhong Guo},
  \bibinfo{person}{Jiaqi Ni}, \bibinfo{person}{Jiashi Li},
  \bibinfo{person}{Jiawei Wang}, \bibinfo{person}{Jin Chen},
  \bibinfo{person}{Jingchang Chen}, \bibinfo{person}{Jingyang Yuan},
  \bibinfo{person}{Junjie Qiu}, \bibinfo{person}{Junlong Li},
  \bibinfo{person}{Junxiao Song}, \bibinfo{person}{Kai Dong},
  \bibinfo{person}{Kai Hu}, \bibinfo{person}{Kaige Gao}, \bibinfo{person}{Kang
  Guan}, \bibinfo{person}{Kexin Huang}, \bibinfo{person}{Kuai Yu},
  \bibinfo{person}{Lean Wang}, \bibinfo{person}{Lecong Zhang},
  \bibinfo{person}{Lei Xu}, \bibinfo{person}{Leyi Xia}, \bibinfo{person}{Liang
  Zhao}, \bibinfo{person}{Litong Wang}, \bibinfo{person}{Liyue Zhang},
  \bibinfo{person}{Meng Li}, \bibinfo{person}{Miaojun Wang},
  \bibinfo{person}{Mingchuan Zhang}, \bibinfo{person}{Minghua Zhang},
  \bibinfo{person}{Minghui Tang}, \bibinfo{person}{Mingming Li},
  \bibinfo{person}{Ning Tian}, \bibinfo{person}{Panpan Huang},
  \bibinfo{person}{Peiyi Wang}, \bibinfo{person}{Peng Zhang},
  \bibinfo{person}{Qiancheng Wang}, \bibinfo{person}{Qihao Zhu},
  \bibinfo{person}{Qinyu Chen}, \bibinfo{person}{Qiushi Du},
  \bibinfo{person}{R.~J. Chen}, \bibinfo{person}{R.~L. Jin},
  \bibinfo{person}{Ruiqi Ge}, \bibinfo{person}{Ruisong Zhang},
  \bibinfo{person}{Ruizhe Pan}, \bibinfo{person}{Runji Wang},
  \bibinfo{person}{Runxin Xu}, \bibinfo{person}{Ruoyu Zhang},
  \bibinfo{person}{Ruyi Chen}, \bibinfo{person}{S.~S. Li},
  \bibinfo{person}{Shanghao Lu}, \bibinfo{person}{Shangyan Zhou},
  \bibinfo{person}{Shanhuang Chen}, \bibinfo{person}{Shaoqing Wu},
  \bibinfo{person}{Shengfeng Ye}, \bibinfo{person}{Shengfeng Ye},
  \bibinfo{person}{Shirong Ma}, \bibinfo{person}{Shiyu Wang},
  \bibinfo{person}{Shuang Zhou}, \bibinfo{person}{Shuiping Yu},
  \bibinfo{person}{Shunfeng Zhou}, \bibinfo{person}{Shuting Pan},
  \bibinfo{person}{T. Wang}, \bibinfo{person}{Tao Yun}, \bibinfo{person}{Tian
  Pei}, \bibinfo{person}{Tianyu Sun}, \bibinfo{person}{W.~L. Xiao},
  \bibinfo{person}{Wangding Zeng}, \bibinfo{person}{Wanjia Zhao},
  \bibinfo{person}{Wei An}, \bibinfo{person}{Wen Liu}, \bibinfo{person}{Wenfeng
  Liang}, \bibinfo{person}{Wenjun Gao}, \bibinfo{person}{Wenqin Yu},
  \bibinfo{person}{Wentao Zhang}, \bibinfo{person}{X.~Q. Li},
  \bibinfo{person}{Xiangyue Jin}, \bibinfo{person}{Xianzu Wang},
  \bibinfo{person}{Xiao Bi}, \bibinfo{person}{Xiaodong Liu},
  \bibinfo{person}{Xiaohan Wang}, \bibinfo{person}{Xiaojin Shen},
  \bibinfo{person}{Xiaokang Chen}, \bibinfo{person}{Xiaokang Zhang},
  \bibinfo{person}{Xiaosha Chen}, \bibinfo{person}{Xiaotao Nie},
  \bibinfo{person}{Xiaowen Sun}, \bibinfo{person}{Xiaoxiang Wang},
  \bibinfo{person}{Xin Cheng}, \bibinfo{person}{Xin Liu}, \bibinfo{person}{Xin
  Xie}, \bibinfo{person}{Xingchao Liu}, \bibinfo{person}{Xingkai Yu},
  \bibinfo{person}{Xinnan Song}, \bibinfo{person}{Xinxia Shan},
  \bibinfo{person}{Xinyi Zhou}, \bibinfo{person}{Xinyu Yang},
  \bibinfo{person}{Xinyuan Li}, \bibinfo{person}{Xuecheng Su},
  \bibinfo{person}{Xuheng Lin}, \bibinfo{person}{Y.~K. Li},
  \bibinfo{person}{Y.~Q. Wang}, \bibinfo{person}{Y.~X. Wei},
  \bibinfo{person}{Y.~X. Zhu}, \bibinfo{person}{Yang Zhang},
  \bibinfo{person}{Yanhong Xu}, \bibinfo{person}{Yanhong Xu},
  \bibinfo{person}{Yanping Huang}, \bibinfo{person}{Yao Li},
  \bibinfo{person}{Yao Zhao}, \bibinfo{person}{Yaofeng Sun},
  \bibinfo{person}{Yaohui Li}, \bibinfo{person}{Yaohui Wang},
  \bibinfo{person}{Yi Yu}, \bibinfo{person}{Yi Zheng}, \bibinfo{person}{Yichao
  Zhang}, \bibinfo{person}{Yifan Shi}, \bibinfo{person}{Yiliang Xiong},
  \bibinfo{person}{Ying He}, \bibinfo{person}{Ying Tang},
  \bibinfo{person}{Yishi Piao}, \bibinfo{person}{Yisong Wang},
  \bibinfo{person}{Yixuan Tan}, \bibinfo{person}{Yiyang Ma},
  \bibinfo{person}{Yiyuan Liu}, \bibinfo{person}{Yongqiang Guo},
  \bibinfo{person}{Yu Wu}, \bibinfo{person}{Yuan Ou}, \bibinfo{person}{Yuchen
  Zhu}, \bibinfo{person}{Yuduan Wang}, \bibinfo{person}{Yue Gong},
  \bibinfo{person}{Yuheng Zou}, \bibinfo{person}{Yujia He},
  \bibinfo{person}{Yukun Zha}, \bibinfo{person}{Yunfan Xiong},
  \bibinfo{person}{Yunxian Ma}, \bibinfo{person}{Yuting Yan},
  \bibinfo{person}{Yuxiang Luo}, \bibinfo{person}{Yuxiang You},
  \bibinfo{person}{Yuxuan Liu}, \bibinfo{person}{Yuyang Zhou},
  \bibinfo{person}{Z.~F. Wu}, \bibinfo{person}{Z.~Z. Ren},
  \bibinfo{person}{Zehui Ren}, \bibinfo{person}{Zhangli Sha},
  \bibinfo{person}{Zhe Fu}, \bibinfo{person}{Zhean Xu}, \bibinfo{person}{Zhen
  Huang}, \bibinfo{person}{Zhen Zhang}, \bibinfo{person}{Zhenda Xie},
  \bibinfo{person}{Zhengyan Zhang}, \bibinfo{person}{Zhewen Hao},
  \bibinfo{person}{Zhibin Gou}, \bibinfo{person}{Zhicheng Ma},
  \bibinfo{person}{Zhigang Yan}, \bibinfo{person}{Zhihong Shao},
  \bibinfo{person}{Zhipeng Xu}, \bibinfo{person}{Zhiyu Wu},
  \bibinfo{person}{Zhongyu Zhang}, \bibinfo{person}{Zhuoshu Li},
  \bibinfo{person}{Zihui Gu}, \bibinfo{person}{Zijia Zhu},
  \bibinfo{person}{Zijun Liu}, \bibinfo{person}{Zilin Li},
  \bibinfo{person}{Ziwei Xie}, \bibinfo{person}{Ziyang Song},
  \bibinfo{person}{Ziyi Gao}, {and} \bibinfo{person}{Zizheng Pan}.}
  \bibinfo{year}{2025}\natexlab{b}.
\newblock \bibinfo{title}{{DeepSeek}-{V3} {Technical} {Report}}.
\newblock
\href{https://doi.org/10.48550/arXiv.2412.19437}{doi:\nolinkurl{10.48550/arXiv.2412.19437}}
\newblock
\shownote{arXiv:2412.19437 [cs]}.


\bibitem[Ferguson et~al\mbox{.}(2012)]%
        {ferguson2012jockey}
\bibfield{author}{\bibinfo{person}{Andrew~D Ferguson}, \bibinfo{person}{Peter
  Bodik}, \bibinfo{person}{Srikanth Kandula}, \bibinfo{person}{Eric Boutin},
  {and} \bibinfo{person}{Rodrigo Fonseca}.} \bibinfo{year}{2012}\natexlab{}.
\newblock \showarticletitle{Jockey: guaranteed job latency in data parallel
  clusters}. In \bibinfo{booktitle}{\emph{Proceedings of the 7th ACM european
  conference on Computer Systems}}. \bibinfo{pages}{99--112}.
\newblock


\bibitem[Gim et~al\mbox{.}(2024)]%
        {gim_promptcache_2024}
\bibfield{author}{\bibinfo{person}{In Gim}, \bibinfo{person}{Guojun Chen},
  \bibinfo{person}{Seung-seob Lee}, \bibinfo{person}{Nikhil Sarda},
  \bibinfo{person}{Anurag Khandelwal}, {and} \bibinfo{person}{Lin Zhong}.}
  \bibinfo{year}{2024}\natexlab{}.
\newblock \showarticletitle{Prompt Cache: Modular Attention Reuse for
  Low-Latency Inference}. In \bibinfo{booktitle}{\emph{Proceedings of Machine
  Learning and Systems}}, Vol.~\bibinfo{volume}{6}.
\newblock


\bibitem[Kwon et~al\mbox{.}(2023a)]%
        {kwon2023vllm}
\bibfield{author}{\bibinfo{person}{Woosuk Kwon}, \bibinfo{person}{Zhuohan Li},
  \bibinfo{person}{Siyuan Zhuang}, \bibinfo{person}{Ying Sheng},
  \bibinfo{person}{Lianmin Zheng}, \bibinfo{person}{Cody~Hao Yu},
  \bibinfo{person}{Joseph~E. Gonzalez}, \bibinfo{person}{Hao Zhang}, {and}
  \bibinfo{person}{Ion Stoica}.} \bibinfo{year}{2023}\natexlab{a}.
\newblock \showarticletitle{Efficient Memory Management for Large Language
  Model Serving with {PagedAttention}}. In
  \bibinfo{booktitle}{\emph{Proceedings of the ACM SIGOPS 29th Symposium on
  Operating Systems Principles (SOSP)}}.
\newblock


\bibitem[Kwon et~al\mbox{.}(2023b)]%
        {kwon_vllm_2023}
\bibfield{author}{\bibinfo{person}{Woosuk Kwon}, \bibinfo{person}{Zhuohan Li},
  \bibinfo{person}{Siyuan Zhuang}, \bibinfo{person}{Ying Sheng},
  \bibinfo{person}{Lianmin Zheng}, \bibinfo{person}{Cody~Hao Yu},
  \bibinfo{person}{Joseph~E. Gonzalez}, \bibinfo{person}{Hao Zhang}, {and}
  \bibinfo{person}{Ion Stoica}.} \bibinfo{year}{2023}\natexlab{b}.
\newblock \showarticletitle{Efficient Memory Management for Large Language
  Model Serving with PagedAttention}. In \bibinfo{booktitle}{\emph{Proceedings
  of the 29th Symposium on Operating Systems Principles}}.
  \bibinfo{pages}{611--626}.
\newblock


\bibitem[Liu et~al\mbox{.}(2025)]%
        {liu2025lmcache}
\bibfield{author}{\bibinfo{person}{Yuhan Liu}, \bibinfo{person}{Yihua Cheng},
  \bibinfo{person}{Jiayi Yao}, \bibinfo{person}{Yuwei An},
  \bibinfo{person}{Xiaokun Chen}, \bibinfo{person}{Shaoting Feng},
  \bibinfo{person}{Yuyang Huang}, \bibinfo{person}{Samuel Shen},
  \bibinfo{person}{Rui Zhang}, \bibinfo{person}{Kuntai Du}, {et~al\mbox{.}}}
  \bibinfo{year}{2025}\natexlab{}.
\newblock \showarticletitle{Lmcache: An efficient KV cache layer for
  enterprise-scale LLM inference}.
\newblock \bibinfo{journal}{\emph{arXiv preprint arXiv:2510.09665}}
  (\bibinfo{year}{2025}).
\newblock


\bibitem[Megiddo and Modha(2003)]%
        {ARC}
\bibfield{author}{\bibinfo{person}{Nimrod Megiddo} {and}
  \bibinfo{person}{Dharmendra~S. Modha}.} \bibinfo{year}{2003}\natexlab{}.
\newblock \showarticletitle{{ARC:} {A} Self-Tuning, Low Overhead Replacement
  Cache}. In \bibinfo{booktitle}{\emph{Proceedings of the {FAST} '03 Conference
  on File and Storage Technologies, March 31 - April 2, 2003, Cathedral Hill
  Hotel, San Francisco, California, {USA}}},
  \bibfield{editor}{\bibinfo{person}{Jeff Chase}} (Ed.).
  \bibinfo{publisher}{{USENIX}}.
\newblock
\urldef\tempurl%
\url{http://www.usenix.org/events/fast03/tech/megiddo.html}
\showURL{%
\tempurl}


\bibitem[{MiniMax}(2026)]%
        {minimax_m25_2026}
\bibfield{author}{\bibinfo{person}{{MiniMax}}.}
  \bibinfo{year}{2026}\natexlab{}.
\newblock \bibinfo{title}{{MiniMax M2.5: Built for Real-World Productivity}}.
\newblock
  \bibinfo{howpublished}{\url{https://www.minimax.io/news/minimax-m25}}.
\newblock
\newblock
\shownote{Accessed: 2026-05-15}.


\bibitem[Patel et~al\mbox{.}(2024)]%
        {patel_splitwise_2024}
\bibfield{author}{\bibinfo{person}{Pratyush Patel}, \bibinfo{person}{Esha
  Choukse}, \bibinfo{person}{Chaojie Zhang}, \bibinfo{person}{Aashaka Shah},
  \bibinfo{person}{{\'I}{\~n}igo Goiri}, \bibinfo{person}{Saeed Maleki}, {and}
  \bibinfo{person}{Ricardo Bianchini}.} \bibinfo{year}{2024}\natexlab{}.
\newblock \showarticletitle{Splitwise: Efficient Generative LLM Inference Using
  Phase Splitting}. In \bibinfo{booktitle}{\emph{Proceedings of the 51st Annual
  International Symposium on Computer Architecture}}.
\newblock


\bibitem[Qin et~al\mbox{.}(2025)]%
        {qin_mooncake_2025}
\bibfield{author}{\bibinfo{person}{Ruoyu Qin}, \bibinfo{person}{Zheming Li},
  \bibinfo{person}{Weiran He}, \bibinfo{person}{Mingxing Zhang},
  \bibinfo{person}{Yongwei Wu}, \bibinfo{person}{Weimin Zheng}, {and}
  \bibinfo{person}{Xinran Xu}.} \bibinfo{year}{2025}\natexlab{}.
\newblock \showarticletitle{Mooncake: A KVCache-Centric Disaggregated
  Architecture for LLM Serving}. In \bibinfo{booktitle}{\emph{23rd USENIX
  Conference on File and Storage Technologies (FAST 25)}}.
  \bibinfo{pages}{155--170}.
\newblock


\bibitem[Qiu et~al\mbox{.}(2025)]%
        {qiu_modserve_2025}
\bibfield{author}{\bibinfo{person}{Haoran Qiu}, \bibinfo{person}{Anish Biswas},
  \bibinfo{person}{Zihan Zhao}, \bibinfo{person}{Jayashree Mohan},
  \bibinfo{person}{Alind Khare}, \bibinfo{person}{Esha Choukse},
  \bibinfo{person}{Íñigo Goiri}, \bibinfo{person}{Zeyu Zhang},
  \bibinfo{person}{Haiying Shen}, \bibinfo{person}{Chetan Bansal},
  \bibinfo{person}{Ramachandran Ramjee}, {and} \bibinfo{person}{Rodrigo
  Fonseca}.} \bibinfo{year}{2025}\natexlab{}.
\newblock \bibinfo{title}{{ModServe}: {Scalable} and {Resource}-{Efficient}
  {Large} {Multimodal} {Model} {Serving}}.
\newblock
\href{https://doi.org/10.48550/arXiv.2502.00937}{doi:\nolinkurl{10.48550/arXiv.2502.00937}}
\newblock
\shownote{arXiv:2502.00937 [cs]}.


\bibitem[Song et~al\mbox{.}(2020)]%
        {LRB}
\bibfield{author}{\bibinfo{person}{Zhenyu Song}, \bibinfo{person}{Daniel~S.
  Berger}, \bibinfo{person}{Kai Li}, {and} \bibinfo{person}{Wyatt Lloyd}.}
  \bibinfo{year}{2020}\natexlab{}.
\newblock \showarticletitle{Learning Relaxed Belady for Content Distribution
  Network Caching}. In \bibinfo{booktitle}{\emph{17th {USENIX} Symposium on
  Networked Systems Design and Implementation, {NSDI} 2020, Santa Clara, CA,
  USA, February 25-27, 2020}}, \bibfield{editor}{\bibinfo{person}{Ranjita
  Bhagwan} {and} \bibinfo{person}{George Porter}} (Eds.).
  \bibinfo{publisher}{{USENIX} Association}, \bibinfo{pages}{529--544}.
\newblock
\urldef\tempurl%
\url{https://www.usenix.org/conference/nsdi20/presentation/song}
\showURL{%
\tempurl}


\bibitem[Srivatsa et~al\mbox{.}(2025)]%
        {srivatsa_preble_2025}
\bibfield{author}{\bibinfo{person}{Vikranth Srivatsa}, \bibinfo{person}{Zijian
  He}, \bibinfo{person}{Reyna Abhyankar}, \bibinfo{person}{Dongming Li}, {and}
  \bibinfo{person}{Yiying Zhang}.} \bibinfo{year}{2025}\natexlab{}.
\newblock \showarticletitle{Efficient Distributed Prompt Scheduling for LLM
  Serving}. In \bibinfo{booktitle}{\emph{International Conference on Learning
  Representations}}.
\newblock


\bibitem[Wang et~al\mbox{.}(2025c)]%
        {wang2025kvcache}
\bibfield{author}{\bibinfo{person}{Jiahao Wang}, \bibinfo{person}{Jinbo Han},
  \bibinfo{person}{Xingda Wei}, \bibinfo{person}{Sijie Shen},
  \bibinfo{person}{Dingyan Zhang}, \bibinfo{person}{Chenguang Fang},
  \bibinfo{person}{Rong Chen}, \bibinfo{person}{Wenyuan Yu}, {and}
  \bibinfo{person}{Haibo Chen}.} \bibinfo{year}{2025}\natexlab{c}.
\newblock \showarticletitle{{KVCache Cache} in the Wild: Characterizing and
  Optimizing {KVCache Cache} at a Large Cloud Provider}. In
  \bibinfo{booktitle}{\emph{2025 USENIX Annual Technical Conference (USENIX ATC
  25)}}. \bibinfo{publisher}{USENIX Association}, \bibinfo{address}{Boston,
  MA}.
\newblock


\bibitem[Wang et~al\mbox{.}(2025d)]%
        {wang_kvcache_2025}
\bibfield{author}{\bibinfo{person}{Jiahao Wang}, \bibinfo{person}{Jinbo Han},
  \bibinfo{person}{Xingda Wei}, \bibinfo{person}{Sijie Shen},
  \bibinfo{person}{Dingyan Zhang}, \bibinfo{person}{Chenguang Fang},
  \bibinfo{person}{Rong Chen}, \bibinfo{person}{Wenyuan Yu}, {and}
  \bibinfo{person}{Haibo Chen}.} \bibinfo{year}{2025}\natexlab{d}.
\newblock \showarticletitle{{KVCache} {Cache} in the {Wild}: {Characterizing}
  and {Optimizing} {KVCache} {Cache} at a {Large} {Cloud} {Provider}}.
  \bibinfo{pages}{465--482}.
\newblock
\showISBNx{978-1-939133-48-9}
\urldef\tempurl%
\url{https://www.usenix.org/conference/atc25/presentation/wang-jiahao}
\showURL{%
\tempurl}


\bibitem[Wang et~al\mbox{.}(2025a)]%
        {wang2025burstgpt}
\bibfield{author}{\bibinfo{person}{Yuxin Wang}, \bibinfo{person}{Yuhan Chen},
  \bibinfo{person}{Zeyu Li}, \bibinfo{person}{Xueze Kang},
  \bibinfo{person}{Yuchu Fang}, \bibinfo{person}{Yeju Zhou},
  \bibinfo{person}{Yang Zheng}, \bibinfo{person}{Zhenheng Tang},
  \bibinfo{person}{Xin He}, \bibinfo{person}{Rui Guo}, \bibinfo{person}{Xin
  Wang}, \bibinfo{person}{Qiang Wang}, \bibinfo{person}{Amelie~Chi Zhou}, {and}
  \bibinfo{person}{Xiaowen Chu}.} \bibinfo{year}{2025}\natexlab{a}.
\newblock \showarticletitle{{BurstGPT}: A Real-World Workload Dataset to
  Optimize LLM Serving Systems}. In \bibinfo{booktitle}{\emph{Proceedings of
  the 31st ACM SIGKDD Conference on Knowledge Discovery and Data Mining V.2
  (KDD '25)}}. \bibinfo{publisher}{ACM}, \bibinfo{address}{Toronto, ON,
  Canada}.
\newblock
\href{https://doi.org/10.1145/3711896.3737413}{doi:\nolinkurl{10.1145/3711896.3737413}}


\bibitem[Wang et~al\mbox{.}(2025b)]%
        {wang_burstgpt_2025}
\bibfield{author}{\bibinfo{person}{Yuxin Wang}, \bibinfo{person}{Yuhan Chen},
  \bibinfo{person}{Zeyu Li}, \bibinfo{person}{Xueze Kang},
  \bibinfo{person}{Yuchu Fang}, \bibinfo{person}{Yeju Zhou},
  \bibinfo{person}{Yang Zheng}, \bibinfo{person}{Zhenheng Tang},
  \bibinfo{person}{Xin He}, \bibinfo{person}{Rui Guo}, \bibinfo{person}{Xin
  Wang}, \bibinfo{person}{Qiang Wang}, \bibinfo{person}{Amelie~Chi Zhou}, {and}
  \bibinfo{person}{Xiaowen Chu}.} \bibinfo{year}{2025}\natexlab{b}.
\newblock \bibinfo{title}{{BurstGPT}: {A} {Real}-world {Workload} {Dataset} to
  {Optimize} {LLM} {Serving} {Systems}}.
\newblock
\href{https://doi.org/10.48550/arXiv.2401.17644}{doi:\nolinkurl{10.48550/arXiv.2401.17644}}
\newblock
\shownote{arXiv:2401.17644 [cs] version: 5}.


\bibitem[Xiang et~al\mbox{.}(2025)]%
        {xiang_servegen_2025}
\bibfield{author}{\bibinfo{person}{Yuxing Xiang}, \bibinfo{person}{Xue Li},
  \bibinfo{person}{Kun Qian}, \bibinfo{person}{Wenyuan Yu},
  \bibinfo{person}{Ennan Zhai}, {and} \bibinfo{person}{Xin Jin}.}
  \bibinfo{year}{2025}\natexlab{}.
\newblock \bibinfo{title}{{ServeGen}: {Workload} {Characterization} and
  {Generation} of {Large} {Language} {Model} {Serving} in {Production}}.
\newblock
\href{https://doi.org/10.48550/arXiv.2505.09999}{doi:\nolinkurl{10.48550/arXiv.2505.09999}}
\newblock
\shownote{arXiv:2505.09999 [cs]}.


\bibitem[Yang et~al\mbox{.}(2017)]%
        {yang_mithril_2017}
\bibfield{author}{\bibinfo{person}{Juncheng Yang}, \bibinfo{person}{Reza
  Karimi}, \bibinfo{person}{Trausti Sæmundsson}, \bibinfo{person}{Avani
  Wildani}, {and} \bibinfo{person}{Ymir Vigfusson}.}
  \bibinfo{year}{2017}\natexlab{}.
\newblock \showarticletitle{Mithril: mining sporadic associations for cache
  prefetching}. In \bibinfo{booktitle}{\emph{Proceedings of the 2017
  {Symposium} on {Cloud} {Computing}}} \emph{(\bibinfo{series}{{SoCC}'17})}.
  \bibinfo{publisher}{Association for Computing Machinery},
  \bibinfo{address}{New York, NY, USA}, \bibinfo{pages}{66--79}.
\newblock
\showISBNx{978-1-4503-5028-0}
\href{https://doi.org/10.1145/3127479.3131210}{doi:\nolinkurl{10.1145/3127479.3131210}}


\bibitem[Yang et~al\mbox{.}(2023)]%
        {s3fifo}
\bibfield{author}{\bibinfo{person}{Juncheng Yang}, \bibinfo{person}{Yazhuo
  Zhang}, \bibinfo{person}{Ziyue Qiu}, \bibinfo{person}{Yao Yue}, {and}
  \bibinfo{person}{Rashmi Vinayak}.} \bibinfo{year}{2023}\natexlab{}.
\newblock \showarticletitle{FIFO queues are all you need for cache eviction}.
  In \bibinfo{booktitle}{\emph{Proceedings of the 29th Symposium on Operating
  Systems Principles}} (Koblenz, Germany) \emph{(\bibinfo{series}{SOSP '23})}.
  \bibinfo{publisher}{Association for Computing Machinery},
  \bibinfo{address}{New York, NY, USA}, \bibinfo{pages}{130–149}.
\newblock
\showISBNx{9798400702297}
\href{https://doi.org/10.1145/3600006.3613147}{doi:\nolinkurl{10.1145/3600006.3613147}}


\bibitem[Yao et~al\mbox{.}(2025)]%
        {yao_cacheblend_2025}
\bibfield{author}{\bibinfo{person}{Jiayi Yao}, \bibinfo{person}{Hanchen Li},
  \bibinfo{person}{Yuhan Liu}, \bibinfo{person}{Siddhant Ray},
  \bibinfo{person}{Yihua Cheng}, \bibinfo{person}{Qizheng Zhang},
  \bibinfo{person}{Kuntai Du}, \bibinfo{person}{Shan Lu}, {and}
  \bibinfo{person}{Junchen Jiang}.} \bibinfo{year}{2025}\natexlab{}.
\newblock \showarticletitle{CacheBlend: Fast Large Language Model Serving for
  RAG with Cached Knowledge Fusion}. In \bibinfo{booktitle}{\emph{Proceedings
  of the Twentieth European Conference on Computer Systems}}.
\newblock


\bibitem[Yu et~al\mbox{.}(2022a)]%
        {yu2022orca}
\bibfield{author}{\bibinfo{person}{Gyeong-In Yu}, \bibinfo{person}{Joo~Seong
  Jeong}, \bibinfo{person}{Geon-Woo Kim}, \bibinfo{person}{Soojeong Kim}, {and}
  \bibinfo{person}{Byung-Gon Chun}.} \bibinfo{year}{2022}\natexlab{a}.
\newblock \showarticletitle{Orca: A Distributed Serving System for
  {Transformer-Based} Generative Models}. In \bibinfo{booktitle}{\emph{16th
  USENIX Symposium on Operating Systems Design and Implementation (OSDI 22)}}.
  \bibinfo{publisher}{USENIX Association}, \bibinfo{address}{Carlsbad, CA},
  \bibinfo{pages}{521--538}.
\newblock


\bibitem[Yu et~al\mbox{.}(2022b)]%
        {yu_orca_2022}
\bibfield{author}{\bibinfo{person}{Gyeong-In Yu}, \bibinfo{person}{Joo~Seong
  Jeong}, \bibinfo{person}{Geon-Woo Kim}, \bibinfo{person}{Soojeong Kim}, {and}
  \bibinfo{person}{Byung-Gon Chun}.} \bibinfo{year}{2022}\natexlab{b}.
\newblock \showarticletitle{Orca: A Distributed Serving System for
  Transformer-Based Generative Models}. In \bibinfo{booktitle}{\emph{16th
  USENIX Symposium on Operating Systems Design and Implementation (OSDI 22)}}.
  \bibinfo{pages}{521--538}.
\newblock


\bibitem[Zhang et~al\mbox{.}(2024)]%
        {sieve}
\bibfield{author}{\bibinfo{person}{Yazhuo Zhang}, \bibinfo{person}{Juncheng
  Yang}, \bibinfo{person}{Yao Yue}, \bibinfo{person}{Ymir Vigfusson}, {and}
  \bibinfo{person}{K.V. Rashmi}.} \bibinfo{year}{2024}\natexlab{}.
\newblock \showarticletitle{{SIEVE} is Simpler than {LRU}: an Efficient
  {Turn-Key} Eviction Algorithm for Web Caches}. In
  \bibinfo{booktitle}{\emph{21st USENIX Symposium on Networked Systems Design
  and Implementation (NSDI 24)}}. \bibinfo{publisher}{USENIX Association},
  \bibinfo{address}{Santa Clara, CA}, \bibinfo{pages}{1229--1246}.
\newblock
\showISBNx{978-1-939133-39-7}
\urldef\tempurl%
\url{https://www.usenix.org/conference/nsdi24/presentation/zhang-yazhuo}
\showURL{%
\tempurl}


\bibitem[Zhang et~al\mbox{.}(2023)]%
        {zhang_h2o_2023}
\bibfield{author}{\bibinfo{person}{Zhenyu Zhang}, \bibinfo{person}{Ying Sheng},
  \bibinfo{person}{Tianyi Zhou}, \bibinfo{person}{Tianlong Chen},
  \bibinfo{person}{Lianmin Zheng}, \bibinfo{person}{Ruisi Cai},
  \bibinfo{person}{Zhao Song}, \bibinfo{person}{Yuandong Tian},
  \bibinfo{person}{Christopher R{\'e}}, \bibinfo{person}{Clark Barrett},
  \bibinfo{person}{Zhangyang Wang}, {and} \bibinfo{person}{Beidi Chen}.}
  \bibinfo{year}{2023}\natexlab{}.
\newblock \showarticletitle{H$_2$O: Heavy-Hitter Oracle for Efficient
  Generative Inference of Large Language Models}. In
  \bibinfo{booktitle}{\emph{Advances in Neural Information Processing
  Systems}}, Vol.~\bibinfo{volume}{36}.
\newblock


\bibitem[Zhao et~al\mbox{.}(2024)]%
        {zhao_wildchat_2024}
\bibfield{author}{\bibinfo{person}{Wenting Zhao}, \bibinfo{person}{Xiang Ren},
  \bibinfo{person}{Jack Hessel}, \bibinfo{person}{Claire Cardie},
  \bibinfo{person}{Yejin Choi}, {and} \bibinfo{person}{Yuntian Deng}.}
  \bibinfo{year}{2024}\natexlab{}.
\newblock \bibinfo{title}{{WildChat}: {1M} {ChatGPT} {Interaction} {Logs} in
  the {Wild}}.
\newblock
\href{https://doi.org/10.48550/arXiv.2405.01470}{doi:\nolinkurl{10.48550/arXiv.2405.01470}}
\newblock
\shownote{arXiv:2405.01470 [cs]}.


\bibitem[Zheng et~al\mbox{.}(2024a)]%
        {zheng_lmsys-chat-1m_2024}
\bibfield{author}{\bibinfo{person}{Lianmin Zheng}, \bibinfo{person}{Wei-Lin
  Chiang}, \bibinfo{person}{Ying Sheng}, \bibinfo{person}{Tianle Li},
  \bibinfo{person}{Siyuan Zhuang}, \bibinfo{person}{Zhanghao Wu},
  \bibinfo{person}{Yonghao Zhuang}, \bibinfo{person}{Zhuohan Li},
  \bibinfo{person}{Zi Lin}, \bibinfo{person}{Eric~P. Xing},
  \bibinfo{person}{Joseph~E. Gonzalez}, \bibinfo{person}{Ion Stoica}, {and}
  \bibinfo{person}{Hao Zhang}.} \bibinfo{year}{2024}\natexlab{a}.
\newblock \bibinfo{title}{{LMSYS}-{Chat}-{1M}: {A} {Large}-{Scale}
  {Real}-{World} {LLM} {Conversation} {Dataset}}.
\newblock
\href{https://doi.org/10.48550/arXiv.2309.11998}{doi:\nolinkurl{10.48550/arXiv.2309.11998}}
\newblock
\shownote{arXiv:2309.11998 [cs]}.


\bibitem[Zheng et~al\mbox{.}(2024b)]%
        {zheng2024sglang}
\bibfield{author}{\bibinfo{person}{Lianmin Zheng}, \bibinfo{person}{Liangsheng
  Yin}, \bibinfo{person}{Zhiqiang Xie}, \bibinfo{person}{Chuyue Sun},
  \bibinfo{person}{Jeff Huang}, \bibinfo{person}{Cody~Hao Yu},
  \bibinfo{person}{Shiyi Cao}, \bibinfo{person}{Christos Kozyrakis},
  \bibinfo{person}{Ion Stoica}, \bibinfo{person}{Joseph~E. Gonzalez},
  \bibinfo{person}{Clark Barrett}, {and} \bibinfo{person}{Ying Sheng}.}
  \bibinfo{year}{2024}\natexlab{b}.
\newblock \showarticletitle{{SGLang}: Efficient Execution of Structured
  Language Model Programs}. In \bibinfo{booktitle}{\emph{Advances in Neural
  Information Processing Systems (NeurIPS)}}.
\newblock


\bibitem[Zheng et~al\mbox{.}(2024c)]%
        {zheng_sglang_2024}
\bibfield{author}{\bibinfo{person}{Lianmin Zheng}, \bibinfo{person}{Liangsheng
  Yin}, \bibinfo{person}{Zhiqiang Xie}, \bibinfo{person}{Chuyue Sun},
  \bibinfo{person}{Jeff Huang}, \bibinfo{person}{Cody~Hao Yu},
  \bibinfo{person}{Shiyi Cao}, \bibinfo{person}{Christos Kozyrakis},
  \bibinfo{person}{Ion Stoica}, \bibinfo{person}{Joseph~E. Gonzalez},
  \bibinfo{person}{Clark Barrett}, {and} \bibinfo{person}{Ying Sheng}.}
  \bibinfo{year}{2024}\natexlab{c}.
\newblock \showarticletitle{{SGL}ang: Efficient Execution of Structured
  Language Model Programs}. In \bibinfo{booktitle}{\emph{Advances in Neural
  Information Processing Systems}}, Vol.~\bibinfo{volume}{37}.
\newblock


\bibitem[Zhong et~al\mbox{.}(2024a)]%
        {zhong2024distserve}
\bibfield{author}{\bibinfo{person}{Yinmin Zhong}, \bibinfo{person}{Shengyu
  Liu}, \bibinfo{person}{Junda Chen}, \bibinfo{person}{Jianbo Hu},
  \bibinfo{person}{Yibo Zhu}, \bibinfo{person}{Xuanzhe Liu},
  \bibinfo{person}{Xin Jin}, {and} \bibinfo{person}{Hao Zhang}.}
  \bibinfo{year}{2024}\natexlab{a}.
\newblock \showarticletitle{{DistServe}: Disaggregating Prefill and Decoding
  for Goodput-Optimized Large Language Model Serving}. In
  \bibinfo{booktitle}{\emph{18th USENIX Symposium on Operating Systems Design
  and Implementation (OSDI 24)}}. \bibinfo{publisher}{USENIX Association},
  \bibinfo{address}{Santa Clara, CA}.
\newblock


\bibitem[Zhong et~al\mbox{.}(2024b)]%
        {zhong_distserve_2024}
\bibfield{author}{\bibinfo{person}{Yinmin Zhong}, \bibinfo{person}{Shengyu
  Liu}, \bibinfo{person}{Junda Chen}, \bibinfo{person}{Jianbo Hu},
  \bibinfo{person}{Yibo Zhu}, \bibinfo{person}{Xuanzhe Liu},
  \bibinfo{person}{Xin Jin}, {and} \bibinfo{person}{Hao Zhang}.}
  \bibinfo{year}{2024}\natexlab{b}.
\newblock \showarticletitle{DistServe: Disaggregating Prefill and Decoding for
  Goodput-Optimized Large Language Model Serving}. In
  \bibinfo{booktitle}{\emph{18th USENIX Symposium on Operating Systems Design
  and Implementation (OSDI 24)}}. \bibinfo{pages}{193--210}.
\newblock


\end{thebibliography}

\appendix

\end{document}